\documentclass[twoside]{article}

\usepackage[preprint]{aistats2026}

\usepackage[utf8]{inputenc} 
\usepackage[T1]{fontenc}    
\usepackage{hyperref}       
\usepackage{url}            
\usepackage{booktabs}       
\usepackage{amsfonts}       
\usepackage{nicefrac}       
\usepackage{microtype}      
\usepackage{lipsum}		
\usepackage{graphicx}
\usepackage{wrapfig}
\usepackage{subcaption}
\usepackage{booktabs} 
\usepackage{balance} 
\usepackage{enumitem}
\usepackage{natbib}
\usepackage{doi}
\usepackage{tabularx}
\usepackage{bbm}

\usepackage{multirow}
\usepackage{cuted} 
\usepackage{amsmath}
\usepackage{amssymb}
\usepackage{mathtools}
\usepackage{amsthm}
\usepackage{makecell}

\usepackage{algorithm}%
\usepackage{algorithmic}%

\usepackage[table]{xcolor} 
\usepackage{amssymb} 
\definecolor{green}{RGB}{0, 128, 0}
\definecolor{red}{RGB}{255, 0, 0}

\newcommand{\cmark}{\textcolor{green}{\textbf{\checkmark}}} 
\newcommand{\xmark}{\textcolor{red}{\textbf{\texttimes}}} 

\usepackage{etoolbox}

\usepackage{mdframed}
\usepackage{stfloats}    

\newmdenv[
  linecolor=black,
  linewidth=1pt,
  backgroundcolor=blue!5,
  roundcorner=5pt,
  skipabove=10pt,
  skipbelow=10pt
]{theobox}

\newmdenv[
  linecolor=green!50!black,
  linewidth=1pt,
  backgroundcolor=green!5,
  roundcorner=5pt,
  skipabove=10pt,
  skipbelow=10pt
]{defbox}

\usepackage{thmtools, thm-restate}
\newtheorem{definition}{Definition}

\newtheorem{proposition}{Proposition}

\begin{document}

\runningtitle{Conformal Convolution and Monte Carlo Meta-learners for Predictive Inference of ITEs}
\runningauthor{Jef Jonkers, Jarne Verhaeghe, Glenn Van Wallendael, Luc Duchateau, Sofie Van Hoecke}
\twocolumn[
\aistatstitle{Conformal Convolution and Monte Carlo Meta-learners for Predictive Inference of Individual Treatment Effects}
\aistatsauthor{Jef Jonkers \And Jarne Verhaeghe  \And  Glenn Van Wallendael} 
\aistatsaddress{IDLab \\ Department of Electronics\\ and Information Systems \\ Ghent University, Belgium \And IDLab \\ Department of Electronics\\ and Information Systems \\ Ghent University - imec, Belgium  \And IDLab \\ Department of Electronics\\ and Information Systems \\ Ghent University - imec, Belgium}
\aistatsauthor{Luc Duchateau \And Sofie Van Hoecke}
\aistatsaddress{Biometrics Research Group \\ Department of Morphology,\\ Imaging, Orthopedics, \\ Rehabilitation and Nutrition \\ Ghent University, Belgium \And IDLab \\ Department of Electronics\\ and Information Systems \\ Ghent University - imec, Belgium}
]



\begin{abstract}
  Generating probabilistic forecasts of potential outcomes and individual treatment effects (ITE) is essential for risk-aware decision-making in domains such as healthcare, policy, marketing, and finance.  We propose two novel methods: the conformal convolution T-learner (CCT) and the conformal Monte Carlo (CMC) meta-learner, that generate full predictive distributions of both potential outcomes and ITEs. Our approaches combine weighted conformal predictive systems with either analytic convolution of potential outcome distributions or Monte Carlo sampling, addressing covariate shift through propensity score weighting. In contrast to other approaches that allow the generation of potential outcome predictive distributions, our approaches are model agnostic, universal, and come with finite-sample guarantees of probabilistic calibration under knowledge of the propensity score. Regarding estimating the ITE distribution, we formally characterize how assumptions about potential outcomes' noise dependency impact distribution validity and establish universal consistency under independence noise assumptions. Experiments on synthetic and semi-synthetic datasets demonstrate that the proposed methods achieve probabilistically calibrated predictive distributions while maintaining narrow prediction intervals and having performant continuous ranked probability scores. Besides probabilistic forecasting performance, we observe significant efficiency gains for the CCT- and CMC meta-learners compared to other conformal approaches that produce prediction intervals for ITE with coverage guarantees.
  
  \textbf{Code}: \url{https://github.com/predict-idlab/cct-cmc}.
\end{abstract}

\section{Introduction}
Modern decision-making requires understanding not just average treatment effects but also complete distributions of potential outcomes. Consider a clinician choosing whether to prescribe a hypertension medication: knowing there is a 60\% chance that the drug reduces systolic blood pressure by 15+ mmHg for an individual patient provides much more actionable information than simply knowing it lowers pressure by 12 mmHg on average for a patient with similar characteristics. Similarly, prediction intervals (PI) showing potential effects ranging from 8-16 mmHg offer more insight than point estimates alone, though they still mask important distributional characteristics. Current machine learning approaches often estimate only conditional average treatment effects (CATE) ~\citep{wagerEstimationInferenceHeterogeneous2018, kunzelMetalearnersEstimatingHeterogeneous2019, nieQuasioracleEstimationHeterogeneous2021, kennedyOptimalDoublyRobust2022, curthNonparametricEstimationHeterogeneous2021}, limiting their applicability in uncertain environments, especially in higher-risk settings such as healthcare~\citep{banerjiClinicalAITools2023, feuerriegel_causal_2024}. Distribution estimation approaches for individual treatment effect (ITE) address this limitation by characterizing the full distribution of potential outcomes, enabling comparisons between potential outcome distributions. This allows clinicians to assess the expected average benefit of a treatment as well as the complete profile of possible responses for each individual patient. Table \ref{tab:example-comparison} compares point, interval, and distributional predictions for a hypertension medication example.

\begin{table*}[htbp]
\centering
\caption{Comparison of ITE Estimation Approaches (Hypertension Medication Example). $ITE = Y(1)-Y(0)$ represents individual treatment effect, $\tau$ is a clinical threshold (e.g., 15\,mmHg reduction).}
\label{tab:example-comparison}
\resizebox{0.8\textwidth}{!}{\begin{tabular}{llll}
\toprule
\textbf{Aspect} & \textbf{Point Prediction} & \textbf{Interval Prediction} & \textbf{Distributional Prediction} \\
\midrule
Output & 
Single estimate: 12\,mmHg & 
90\% PI: 8--16\,mmHg & 
Full CDF $\hat{F}_{\Psi}(y)$ \\
\addlinespace

Threshold Query & 
Cannot quantify $P(ITE \geq 15)$ & 
Bounds contain $15$ but no probability & 
Direct: $P(ITE \geq 15) = 0.60$ \\
\addlinespace

Risk Assessment & 
Hides all variance & 
Masks bimodality/outliers & 
Reveals probability mass at thresholds \\
\addlinespace

\makecell[l]{Utility\\Integration} & 
Only computes $U(12)$ & 
Requires worst-case assumptions & 
Enables $\mathbb{E}[U(ITE)] = \int U(y) dF_{ITE}(y)$ \\
\addlinespace

\makecell[l]{Possible Theoretical\\Guarantee} & 
Consistent for CATE & 
Marginal coverage & 
Probabilistic and quantile calibration \\
\bottomrule
\end{tabular}}
\end{table*}

Reliable distributional estimation for ITEs must address three key challenges: \textbf{(1) Confounding} robustness to covariate shift via propensity score adjustment; \textbf{(2) Universal} asymptotic convergence to the data-generating ITE distribution; and \textbf{(3) Finite sample calibration} even for small $n$. As can be seen in Table 2, existing distributional approaches struggle mostly with the third challenge to provide calibrated predictive distributions for counterfactuals and ITEs. Additionally, none of the existing solutions is model agnostic. Bayesian methods (BART~\citep{chipman_bart_2010, hillBayesianNonparametricModeling2011a}, CMGP~\citep{alaaBayesianInferenceIndividualized2017}, DKLITE~\citep{zhang_learning_2020}) and some deep learning approaches (CEVAE~\citep{louizos_causal_2017}) impose restrictive parametric assumptions limiting universality. While other neural network approaches (GANITE~\citep{yoon_ganite_2018}, NOFLITE~\citep{vanderschueren_noflite_2023}, DiffPO~\citep{ma_diffpo_2024}, FCCN~\citep{zhou_estimating_2022}) allow for the modeling of more flexible predictive distributions, they often produce poorly calibrated distributions and lack finite-sample guarantees of probabilistic calibration. Conformal prediction (CP) approaches (WCP~\citep{leiConformalInferenceCounterfactuals2021}, CM-learners \citep{alaaConformalMetalearnersPredictive2023}) advance interval estimation but cannot quantify threshold probabilities or integrate utility functions. These limitations persist because counterfactual inference fundamentally involves unobserved outcomes, making it a challenge to estimate predictive distributions while balancing robustness, calibration, and computational tractability.

\begin{table*}[htbp]
\centering
\small
\caption{Comparison of different distributional approaches in the literature.}
\label{tab:sota-comparison}
\resizebox{0.8\textwidth}{!}{\begin{tabular}{@{}llccccc@{}}
\toprule
\textbf{Approach} & \textbf{Methods} & \makecell{\textbf{S-, T-, or Direct-} \\ \textbf{learner}}& \makecell{\textbf{Confounding}} & \makecell{\textbf{Universal}} & \makecell{\textbf{Finite}\\\textbf{sample}} & \makecell{\textbf{Model}\\\textbf{agnostic}} \\ 
\midrule
BART  \citep{chipman_bart_2010, hillBayesianNonparametricModeling2011a}              & BART             & S-learner  & \xmark      & \xmark        & \xmark         & \xmark          \\
CEVAE \citep{louizos_causal_2017}            & VAE              & S-learner             & \xmark        & \xmark        & \xmark         & \xmark          \\
CMGP \citep{alaaBayesianInferenceIndividualized2017}             & GP               & S-learner          & \cmark        & \xmark        & \xmark         & \xmark          \\
GANITE \citep{yoon_ganite_2018}           & GAN              & Direct-learner           & \cmark        & \xmark        & \xmark         & \xmark          \\
DKLITE \citep{zhang_learning_2020}           & Deep kernels     & S-learner          & ?             & \xmark        & \xmark         & \xmark          \\
NOFLITE  \citep{vanderschueren_noflite_2023}         & NFs        & T or S        & \cmark        & \xmark        & \xmark         & \xmark          \\
DiffPO  \citep{ma_diffpo_2024}          & DM        & S-learner          & \cmark        & \xmark        & \xmark         & \xmark          \\
FCCN \citep{zhou_estimating_2022}             & CN   & T-learner      & \cmark        & \cmark        & \xmark         & \xmark          \\
\midrule
\textbf{CCT-learner} (ours)     & WCPS   & T-learner    & \cmark        & \cmark        & \cmark         & \cmark          \\
\textbf{CMC Meta-learner} (ours) & WCPS   & T or S      & \cmark        & \cmark        & \cmark         & \cmark          \\ 
\bottomrule
\end{tabular}}
\end{table*}

We introduce 3 advances to make distributional regression for causal machine learning more reliable: 

\paragraph{1. ~Conformalized ITE distribution estimation} We propose a novel model-agnostic framework for deriving nonparametric predictive distributions for counterfactual outcomes and ITEs, based on (weighted) conformal predictive systems (WCPS)\citep{vovkNonparametricPredictiveDistributions2019, jonkers2024conformal}, a framework that applies (weighted) CP (WCP) \citep{vovkAlgorithmicLearningRandom2022, tibshiraniConformalPredictionCovariate2019a} to derive probabilistically calibrated predictive distributions under the (weighted) exchangeability assumption, resulting in two approaches:
\begin{itemize}[leftmargin=*, noitemsep, nosep]
    \item \textbf{CCT-learner} constructs predictive distributions of ITEs via analytical convolution of propensity-weighted conformal predictive distributions from the potential outcome models, which under certain assumptions guarantee probabilistic calibration of the predictive distribution (Theorem \ref{theo:cct-calibration}). 
    \item \textbf{CMC-learner} trades finite-sample ITE calibration and some predictive performance for computational efficiency, using Monte Carlo sampling to approximate distributions while still allowing for universal consistency (Theorem \ref{theo:consistency}).
\end{itemize}

\paragraph{2. ~Theoretical guarantees and results} We establish rigorous theoretical foundations for WCPS and our introduced framework:
\begin{itemize}[leftmargin=*,noitemsep, nosep]
    \item \textbf{Finite-sample calibration} for potential outcome distributions under covariate shift, by proving that WCPS \citep{jonkers2024conformal} are guaranteed to be probabilistically calibrated (Theorem \ref{theo:valid-wct}). This ensures the probabilistic calibration of the potential outcome predictive distributions and allows us to prove that CCT-learners have a finite-sample calibration guarantee under certain assumptions.
    \item \textbf{Existence of universal consistent solution}, proving that the CCT- and the CMC meta-learners converge to the data-generating ITE distribution as $n\rightarrow \infty$, assuming the underlying potential outcome models converge (Theorem \ref{theo:consistency}).
    \item \textbf{Noise dependence characterization}, formalizing how correlations between noise on potential outcomes affect validity, a critical and often neglected factor in ITE and CATE estimation, especially for uncertainty quantification. 
\end{itemize}

\paragraph{3. ~Empirical robustness} Our methods outperform or are on par with existing approaches for prediction interval and predictive distribution estimation across six synthetic and three semi-synthetic datasets, with a total of 88 different settings. The extensive evaluation demonstrates that our methods:
\begin{itemize}[leftmargin=*,noitemsep, nosep]
\item Maintain \textbf{probabilistic calibration} even under model misspecification and under propensity score estimation, ensuring reliability in uncertainty quantification across diverse data scenarios.
\item Show superior performance related to continuous ranked probability score (CRPS) across multiple datasets and settings, indicating more accurate predictive distributions than SOTA.
\item Deliver consistent coverage properties in prediction intervals, with empirical coverage rates closely matching nominal coverage levels even in challenging scenarios.
\item Exhibit exceptional \textbf{flexibility and model-agnosticity}, seamlessly integrating with various causal inference models and distribution estimation approaches, highlighting the versatility of our method across different modeling paradigms.
\end{itemize}

Code and datasets are available at \url{https://github.com/predict-idlab/cct-cmc}.
Code and datasets are available in the supplementary material.

\section{Problem definition}
\label{sec:prob-def}
In this work, we use the Neyman-Rubin~\citep{neyman_application_1990, rubinCausalInferenceUsing2005} potential outcome framework to formulate our problem. We assume access to a sample of observations $\mathcal{D} = \{Z_i=(Y_i, X_i, W_i)\}^n_{i=1}$, with $Z_i \overset{\mathrm{iid}}{\sim} \mathcal{P}$ with $\mathcal{P}$ the resulting joint distribution of the data generation process. Here, $Y_i \in \mathcal{Y}$ represents the continuous observed outcome of interest, $X_i \in \mathcal{X} \subset \mathbb{R}^d$ denotes the per-object feature values, and $W_i \in \{0,1\}$ is the binary treatment assignment. The joint distribution $\mathcal{P}$ is defined by the following covariate distribution and nuisance functions:
\setlength{\abovedisplayskip}{3pt}
\setlength{\belowdisplayskip}{0pt}
\begin{align*}
& \pi(x) = \mathbb{P}(W=1|X=x),\quad W \sim \text{Bernoulli}(\pi(x)),\\
& X \sim \Lambda, \quad Y(w) = \mu_w(X) + \varepsilon(w),\\
& \mu_w(x) = \mathbb{E}(Y|X=x,W=w), \\
& Y = Y(0)(1-W) + Y(1)W, \\
& \tau(x) = \mathbb{E}(Y(1)-Y(0)|X=x), \\
& ITE = Y(1) - Y(0) = \tau(X) + \varepsilon_{a, ITE}
\end{align*}
where $\pi(x)$ denotes the propensity score that defines the probability of treatment assignment for an individual, $\mu_w(x)$ represents the expected (potential) outcome function of an individual for treatment assignment $w$, $\Lambda$ represents the marginal distribution of $X$, $Y(w)$ represent the potential outcome when an individual receives treatment $w$, $\varepsilon(w)$ are zero-mean random variables independent of $X$ and $W$, $Y$ is the observed outcome, $\tau(x)$ denotes the CATE function, $ITE$ the individual treatment effect (ITE), and $\varepsilon_{a, ITE}$ represent a zero-mean random variable. \\
Next, we assume that the described data-generating process follows the following assumptions: (1) \textit{Consistency:} $Y(w_i) = Y_i$; (2) \textit{Unconfoundedness:} $Y(0), Y(1) \perp W | X$ and (3) \textit{Positivity:} $Pr[W=w|X=x] > 0$.
Under Assumption (1) and (2), $\mu_w$ becomes a potential outcome as $\mathbb{E}(Y(w)|X) = \mathbb{E}(Y|X=x,W=w)$, aiding in estimating $\tau$ and defining the potential outcome $\mu_w$.\\
In this work, we aim to estimate $ITE_i$, defined as the difference between potential outcomes $Y_i(1)$ and $Y_i(0)$. However, as observing both outcomes for the same individual $i$ is impossible, known as the fundamental problem of causal inference, previous works focused on estimating the CATE function $\tau(x)$~\citep{wagerEstimationInferenceHeterogeneous2018, kunzelMetalearnersEstimatingHeterogeneous2019, nieQuasioracleEstimationHeterogeneous2021, kennedyOptimalDoublyRobust2022, curthNonparametricEstimationHeterogeneous2021}. When conditioning on the same covariate set, the optimal estimator for the CATE function also serves as the optimal estimator for the ITE in terms of mean squared error~\citep{kunzelMetalearnersEstimatingHeterogeneous2019}. However, as already stated, CATE approaches predominantly yield point estimates and often overlook quantifying uncertainty in predictions. While the intrinsic value of CATE estimates is undeniable, quantifying uncertainty surrounding these estimates is essential for robust decision-making in high-risk environments~\citep{banerjiClinicalAITools2023, feuerriegel_causal_2024}. Therefore, it is crucial to address both dimensions of predictive uncertainty, $\widehat{ITE} = \hat{\tau}(X) + \varepsilon_{a, ITE}(X) = \tau(X) + \varepsilon_{a, ITE}(X) + \varepsilon_{e, ITE}(X)$ with the epistemic uncertainty ($\varepsilon_{e, ITE}$) originating from model misspecification and finite samples, and the aleatoric uncertainty ($\varepsilon_{a, ITE} = \varepsilon(1) - \varepsilon(0)$) arising from the inherent variability of ITE among individuals with the same covariates. Incorporating a comprehensive understanding of these uncertainties is pivotal for advancing the reliability and applicability of heterogeneous treatment effect models.

\section{Conformal predictive systems}
\label{sec:cps}
\citet{vovkNonparametricPredictiveDistributions2019, vovkComputationallyEfficientVersions2020} propose conformal predictive systems (CPS), an extension of CP (Appendix \ref{ap:cp}) that allows deriving predictive distributions and enjoys provable properties of validity, in the sense of probabilistic calibration, under exchangeability~\citep{vovkNonparametricPredictiveDistributions2019}. CPS produces conformal predictive distributions as p-values arranged into a probability distribution function~\cite{vovkNonparametricPredictiveDistributions2019}, created with the help of conformity scores. \citet{vovkNonparametricPredictiveDistributions2019} define a CPS as a function that is both a smooth conformal transducer and a randomized predictive system (RPS). An RPS is a function that, given a training dataset, a test input, a candidate label $y$, and a random variable $\phi \sim Uniform(0,1)$, outputs a value in  $[0,1]$ representing the cumulative probability of observing a label less than or equal to $y$. It is required to be monotonic in both $y$ and 
$\phi$ converges to 0 and 1 as $y \rightarrow -\infty$ and $y \rightarrow \infty$, and produces uniformly calibrated predictive distributions, meaning it behaves like a valid CDF and is calibrated under exchangeability. This RPS is used to formalize the theoretical framework of CPS, enabling us to prove that CPS distributional predictions have finite-sample validity. The conformal transducer is another way to represent a conformal predictor, i.e., a confidence transducer that can be seen as a function that maps each input sequence into a sequence of p-values \citep{vovkAlgorithmicLearningRandom2022}. Formally, we define the full smoothed conformal transducer, defined by conformity score $s$, as
\begin{align*}
    &Q(X_{n+1},y,\phi; Z_{1:n}) &:= \sum_{i=1}^{n+1} [S_i^y < S_{n+1}^y] \frac{1}{n+1} + \\&& \sum_{i=1}^{n+1} [S_i^y = S_{n+1}^y] \frac{\phi}{n+1} 
\end{align*}
where $Z_{1:n}$ is the training sequence, $\phi \sim Uniform(0,1)$, $X_{n+1}$ is a test object, and for each label $y$ the corresponding conformity score $S_i^y$ is defined as
\begin{equation*}
\label{eq:CPS}
    \begin{split} 
        S_i^y &:= s((X_i, Y_i); Z_{1:n \setminus i} \cup  (X_{n+1}, y)), \qquad i = 1, ..., n \\
        S_{n+1}^y &:= s((X_{n+1}, y); Z_{1:n}).
    \end{split}
\end{equation*}
An RPS is defined as a function $Q: \mathcal{Z}^{n+1} \times [0,1] \rightarrow [0,1]$ that satisfies three requirements: (1) the function $Q(X_{n+1},y,\phi; Z_{1:n})$ needs to be monotonically increasing both in $y \in \mathbb{R}$ and $\phi \in [0,1]$; (2) $\lim_{y \rightarrow -\infty} Q(X_{n+1},y,0; Z_{1:n}) = 0$ and $\lim_{y \rightarrow \infty} Q(X_{n+1},y,1; Z_{1:n}) = 1$; and (3) it must be probabilistically calibrated,   $\forall \alpha \in [0,1]: \mathbb{P}\{Q(X_{n+1},Y_{n+1},\phi; Z_{1:n}) \leq \alpha\} = \alpha. $ \\
A smoothed conformal transducer is an RPS if and only if the exchangeability assumption is satisfied and if it is defined by a monotonic conformity score \citep{vovkNonparametricPredictiveDistributions2019}.
CPS has been adapted to become more computationally efficient, i.e., split conformal predictive systems (SCPS) \citep{vovkComputationallyEfficientVersions2020}, by building upon inductive CP (ICP). Here, the p-values are created by a split conformity score function that is an element of a family of measurable functions $s: \mathcal{Z} \times \mathcal{Z}^m \rightarrow \mathbb{R} \cup \{-\infty,\infty\}$, where $m$ is the length of the training sequence. This score needs to be isotonic and balanced. Examples of such a measure are the residual error $y-\hat{f}(x)$ and normalized residual error $\frac{y-\hat{f}(x)}{\hat{u}(x)}$, where $\hat{f}(\cdot)$ represent the fitted regression model and $\hat{u}(\cdot)$ a 1D-uncertainty estimate. The SCPS is a function that is both a split-smoothed conformal transducer and an RPS and follows a similar form as in Eq. \ref{eq:CPS}, however, only leveraging conformity scores of the calibration set. \citet{vovkComputationallyEfficientVersions2020} proved that the split smoothed conformal transducer is an RPS if the exchangeability assumption is satisfied and the conformity measure is isotonic and balanced.
In this work, we also introduce a new conformity score for CPS, namely, the probability integral transform (PIT)-values $s=\hat{F}_{X_i}(Y_i)$ of the distribution prediction $\hat{F}_{X_i}$. For this score, you need access to the CDF estimate and thus require a distributional regression approach as a base-learner. Note that \citet{chernozhukov_distributional_2018} introduced a similar score in their distributional CP approach.

\subsection{Conformal predictive systems under covariate shift}
CP and CPS validity rely on the exchangeability assumption. However, in counterfactual inference, except in randomized control trials (RCTs), these assumptions are often violated due to covariate shifts. Treatment assignment probabilities, depending on covariates, cause this shift, affecting the modeling of counterfactual outcomes. \\
Fortunately, \citet{tibshiraniConformalPredictionCovariate2019a} extended CP beyond the exchangeability assumption, proposing weighted CP (WCP), a weighted version of CP to deal with covariate shifts where the likelihood ratio between the training $\mathcal{P}_X$ and test $\tilde{\mathcal{P}}_X$ is known. WCP reweights each nonconformity score by a probability weight $p^w_i$ proportional to the likelihood ratio $w_i = \frac{d\tilde{\mathcal{P}}_X(X_i)}{d\mathcal{P}_X(X_i)}$, where the weights are defined by $p^w_i = \frac{w(X_i)}{\sum^{n+1}_{j=1} w(X_j)}$. \citet{leiConformalInferenceCounterfactuals2021} implemented WCP for counterfactual inference, next to extending it to conformalized quantile regression (CQR) and giving an upper bound to the conservative validity guarantee of WCP. 
For CPS, \citet{jonkers2024conformal} proposed weighted CPS (WCPS), similar to WCP, to allow for probabilistically calibrated predictive distributions under covariate shift. However, while \citet{jonkers2024conformal} proposes the framework for WCPS by introducing a weighted (split) conformal transducer, they omit the proof of the probabilistic calibration of the weighted conformal transducer (Theorem \ref{theo:valid-wct}) and only validate it empirically. We introduce this proof in this work (Appendix \ref{ap:proof-wct}).

\begin{theobox}
\begin{restatable}{theorem}{theoWeightedConformalTransducer}
\label{theo:valid-wct}
Assume that $Z_i = (X_i, Y_i) \in \mathcal{X} \times \mathbb{R}$, $i=1,...,n$ are produced independently from $\mathcal{P} = \mathcal{P}_X \times \mathcal{P}_{Y|X}$; $Z_{n+1} = (X_{n+1}, Y_{n+1}) \in \mathcal{X} \times \mathbb{R}$, is independently drawn from $\tilde{\mathcal{P}} = \tilde{\mathcal{P}}_X \times \mathcal{P}_{Y|X}$; $\tilde{\mathcal{P}}_X$ is absolutely continuous with respect to $\mathcal{P}_X$; likelihood ratio $w(x) = \frac{d\tilde{\mathcal{P}}_X}{d\mathcal{P}_X}$ representing the covariate shift,  random number $\phi \sim Uniform(0,1)$; $Z_{1:n}$, $Z_{n+1}$, and $\phi$ to be independent.
Then, the distribution of the weighted conformal transducer, defined by 
\begin{equation}
    \begin{split}
        Q(X_{n+1},y,\phi, w(x);Z_{1:n}) := \\ \sum_{i=1}^{n+1} [S_i^y < S_{n+1}^y] p_i^w +  \sum_{i=1}^{n+1} [S_i^y = S_{n+1}^y] p_i^w \phi
    \end{split}
\end{equation}
is uniform:
\begin{equation}
    \begin{split}
        \forall \alpha \in [0,1]: \mathbb{P}_{Z_{1:n} \sim \mathcal{P}, (X_{n+1},Y_{n+1}) \sim \tilde{\mathcal{P}}}\{ \\
        Q(X_{n+1},Y_{n+1},\phi, w(x); Z_{1:n}) \leq \alpha\} = \alpha
    \end{split}
\end{equation}
\end{restatable}
\end{theobox}
This result allows us to generate potential outcome predictive distributions, which are guaranteed to be probabilistically calibrated in a finite sample setting under a known covariate shift, for both binary and continuous interventions. Theorem \ref{theo:valid-wct} also applies to split weighted conformal transducer and thus to split WCPS, as this is just a special case of WCPS. In the remaining part of this work, we will work with split WCPS, as it is more computationally efficient and easier to validate whether the score function is isotonic, which is necessary to generate a valid predictive distribution.
Note that the approach of \citet{jonkers2024conformal}  is model-agnostic; by building our framework upon WCPS, it can be used with any point and distributional regression model. We even show in our experiments (Section \ref{sec:exp-results}) that we can wrap WCPS and the in this paper presented distributional meta-learners (Section \ref{sec:method}) around the flexible causal distributional models (FCCN \citep{zhou_estimating_2022} and NOFLITE \citep{vanderschueren_noflite_2023}). In this case, you could view WCPS as a hedge against miscalibration, at the cost of fewer samples to train the model, as some are needed for the calibration set. Additionally, Theorem \ref{theo:valid-wct} also allows us to obtain probabilistically calibrated distributions of individual treatment effects under treatment (ITT) and control (ITC), which can be used as an analysis tool. 

\section{Conformal convolution and conformal Monte Carlo meta-learner}
\label{sec:method}
\begin{figure}[htbp]
    \centering
    \includegraphics[width=\linewidth]{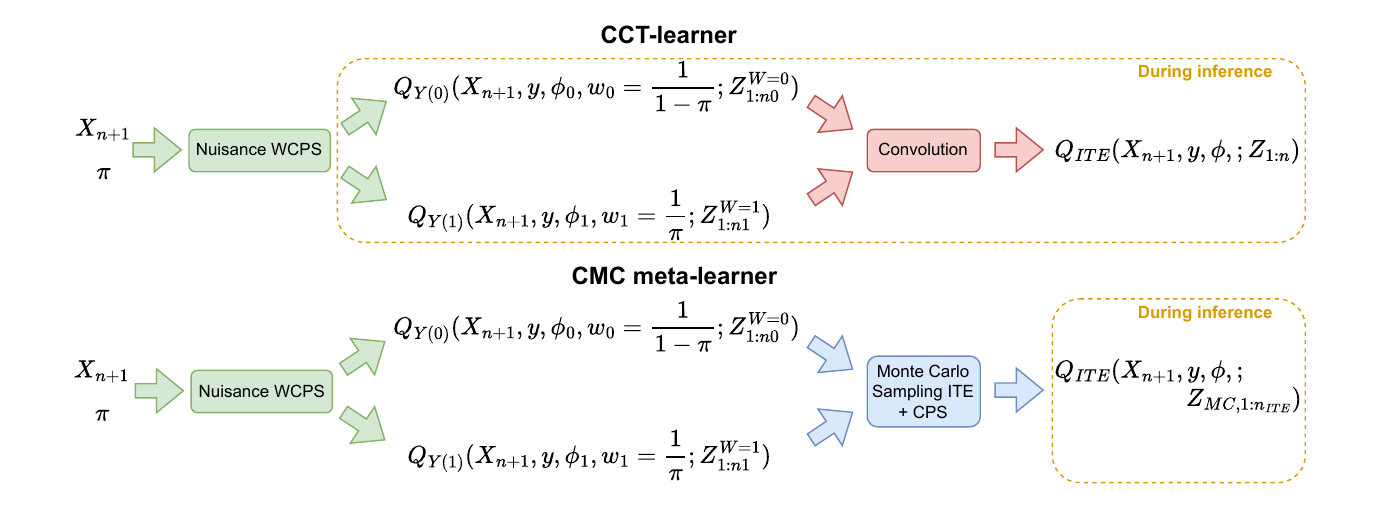}
    \caption{Illustration of CCT- and CMC T-learner algorithms, where $\frac{1}{1-\pi}$ and $\frac{1}{\pi}$ are the proportional likelihood ratio function used to unconfound the outcome under control $Y(0)$ and treatment $Y(1)$, respectively.}
    \label{fig:illustration}
\end{figure}

We introduce the conformal convolution T- (CCT) and conformal Monte Carlo (CMC) meta-learners that both start from counterfactual inference. They use WCPS to generate probabilistically calibrated predictive distributions of the potential outcomes (Section \ref{sec:cps}) to infer the ITE predictive distribution. The CCT- and CMC meta-learner methods are illustrated in Figure \ref{fig:illustration}, with descriptions in Algorithm \ref{alg:fit_cct} and~\ref{alg:fit_cmc}. 

\begin{algorithm}[htbp]
    \caption{Fit and inference of CCT-learner}
    \label{alg:fit_cct}
    \begin{algorithmic}[1]
        \STATE {\bfseries Assume:} $\phi \sim Uniform(0,1)$, $y \in \mathbb{R}$
        \STATE {\bfseries Input:} observations $\{Z_i = (Y_i, X_i, W_i)\}_{i=1}^n$, propensity score function $\pi(x)$, and test object $X_{n+1}$.
        \STATE $Z^{W=0}_{1:n_0} := \{(Y_i, X_i) | W_i = 0\}_{i=1}^n$ where $n_0$ the number observation under control.
        \STATE $Z^{W=1}_{1:n_1} := \{(Y_i, X_i) | W_i = 1\}_{i=1}^n$ where $n_1$ the number observation under treatment.
        \STATE Construct $\textbf{Q}_{Y(0)}(X_{n+1}, y, \phi_0, w_0 = \frac{1}{1-\pi};Z^{W=0}_{1:n_0})$ and $\textbf{Q}_{Y(1)}(X_{n+1}, y, \phi_1, w_1 = \frac{1}{\pi};Z^{W=1}_{1:n_1})$.
        \STATE $\textbf{Q}_{ITE}(X_{n+1}, y, \phi; \textbf{Q}_{Y(0)}, \textbf{Q}_{Y(1)})) = \textbf{Q}_{Y(1)} * \textbf{Q}_{-Y(0)}(X_{n+1}, y, \phi)$ \COMMENT{\textit{Eq. \ref{eq:conv-split}}}
    \end{algorithmic}
\end{algorithm}

\begin{algorithm}[htbp]
    \caption{Fit and inference of CMC meta-learner}
    \label{alg:fit_cmc}
    \begin{algorithmic}[1]
        \STATE {\bfseries Assume:} $\phi \sim Uniform(0,1)$, $y \in \mathbb{R}$
        \STATE {\bfseries Input:} observations $\{Z_i = (Y_i, X_i, W_i)\}_{i=1}^n$, propensity score function $\pi(x)$, test object $X_{n+1}$, and MC samples $n_{MC}$.
        \STATE $Z_{train,1:n_{train}}, Z_{cal,1:n_{cal}} = \textbf{RandomSplit}(Z_{1:n})$
        \STATE $Z^{W=0}_{train,1:n_0} := \{(Y_{train,i}, X_{train,i}) | W_{train,i} = 0\}_{i=1}^{n_{train}}$ where $n_0$ the number observation in the training split under control.
        \STATE $Z^{W=1}_{train,1:n_1} := \{(Y_{train,i}, X_{train,i}) | W_{train,i} = 1\}_{i=1}^{n_{train}}$ where $n_1$ the number observation in the training split under treatment.
        \STATE Construct $\textbf{Q}_{Y(0)}(x, y, \phi_0, w_0 = \frac{1}{1-\pi};Z^{W=0}_{train,1:n_0})$ and $\textbf{Q}_{Y(1)}(x, y, \phi_1, w_1 = \frac{1}{\pi};Z^{W=1}_{train,1:n_1})$.
        \STATE $ITE_{MC,1:n_{ITE}}, X_{MC,1:n_{ITE}} = \textbf{MonteCarloITE}(Z_{cal,1:n_{cal}}, \textbf{Q}_{Y(0)}, \textbf{Q}_{Y(1)}, n_{MC})$ where $n_{ITE} = n_{cal} n_{MC}.$ \COMMENT{\textit{Algorithm \ref{alg:pseudo_MC_ite}}}
        \STATE Construct $\textbf{Q}_{ITE}(X_{n+1}, y, \phi;\{ITE_{MC,i}, X_{MC,i}\}_{i=1}^{n_{ITE}})$.
    \end{algorithmic}
\end{algorithm}

Both approaches deal with covariate shifts using WCPS. These shifts occur because the control and treatment models are trained and calibrated solely on observed outcomes, leading to different covariate distributions between treated and control groups due to confounding variables. To achieve calibrated predictive distributions for outcomes under control $Y(0)$ and treatment $Y(1)$, respectively, we weight each conformity score proportional to $\frac{1}{1-\pi(x)}$ and $\frac{1}{\pi(x)}$ since
\begin{align*}
    w_0(x) &= \frac{d\mathcal{P}_X(x)}{d\mathcal{P}_{X|W=0}(x)} = \frac{d\mathcal{P}_X(x)\mathbb{P}(W=0)}{d\mathcal{P}_X(x)(1-\pi(x))} \propto \frac{1}{1-\pi(x)} \\
    w_1(x) &= \frac{d\mathcal{P}_X(x)}{d\mathcal{P}_{X|W=1}(x)} = \frac{d\mathcal{P}_X(x)\mathbb{P}(W=1)}{d\mathcal{P}_X(x)\pi(x)} \propto \frac{1}{\pi(x)}
\end{align*}
where $\mathcal{P}_X(x)$ denotes the covariate distribution of the entire population, $\mathcal{P}_{X|W=1}(x)$ and $\mathcal{P}_{X|W=0}(x)$ represent the covariate distribution of the treated and control population, respectively, and $\mathbb{P}(W=1)$ and $\mathbb{P}(W=0)$ are the marginal probabilities of receiving treatment and control. The weights match those used by \citet{leiConformalInferenceCounterfactuals2021} for counterfactual inference with WCP and are similar to the inverse probability weighted (IPW) estimation of average causal effects \citep{imbens_causal_2015}.

From WCPS, we infer the potential outcomes predictive distributions, $Q_{-Y(0)}$ and $Q_{Y(1)}$, which are probabilistically calibrated. Since our primary interest often lies in ITE, defined as $ITE = Y(1)-Y(0)$, we introduce a convolution-like operation on the treatment and the negated control predictive distributions, as defined in Eq. \ref{eq:conv-split}. Theorem \ref{theo:cct-calibration} states the finite-sample calibration guarantee of the CCT-learner, and the proof is given in Appendix \ref{ap:proof-cct-validity}.

\begin{figure*}[htbp]  
\centering
\begin{theobox}
\begin{restatable}{theorem}{cctCalibration}
\label{theo:cct-calibration}
Assume that $Z_i = (X_i, W_i, Y_i)  \in \mathcal{X} \times \{0,1\} \times \mathbb{R}$, $i=1,...,n$ and $X_{n+1} \in \mathcal{X}$, $Y_{n+1}(0), Y_{n+1}(1) \in \mathbb{R}$ are produced independently from $\mathcal{P}$, and $\phi \sim Uniform(0,1)$. 

Then, the "convolution" of the distributions $Q_{-Y(0)}$ and $Q_{Y(1)}$, defined by 
\begin{equation}
    \label{eq:conv-split}
    \begin{split}
         Q_{Y(1)-Y(0)}(X_{n+1}, y, \phi, w_0, w_1; Z_{1:n})  &= \sum_{i=1}^{n_0} \sum_{j=1}^{n_1} \left[ \left\{q_{Y(1),j}(X_{n+1}) - q_{Y(0),i}(X_{n+1}) \right\} < y \right] p_{j}^{w_1} p_{i}^{w_0}\\
        &+ \left[ \left\{q_{Y(1),j}(X_{n+1}) - q_{Y(0),i}(X_{n+1}) \right\} = y\right] p_{j}^{w_1} p_{i}^{w_0} \phi \\
        &+ \left[1 - \left\{1-p_{n+1}^{w_1}\right\} \left \{1-p_{n+1}^{w_0} \right\}\right] \phi,
    \end{split}
\end{equation}
where $n_1$ and $n_0$ are the calibration set sizes for the WCPS of outcomes under treatment and control, $q_{Y(1),j}(X_{n+1})$ and $q_{Y(0),i}(X_{n+1})$ are the $j$th and $i$th right-continuous points of $Q_{Y(1)}(X_{n+1}, y, \phi_1=0, w_1 = \frac{1}{\pi};Z^{W=1}_{1:n_1})$ and $Q_{Y(0)}(X_{n+1}, y, \phi_0=0, w_0 = \frac{1}{1-\pi};Z^{W=0}_{1:n_0})$, is uniform:
\begin{equation}
    \begin{split}
        \forall \alpha \in [0,1]: &\mathbb{P}_{Z_{1:n}, X_{n+1}, Y_{n+1}(0), Y_{n+1}(1) \sim \mathcal{P}}\{ \\
        &Q_{Y(1)-Y(0)}(X_{n+1}, Y_{n+1}(1) - Y_{n+1}(0), \phi, w_0, w_1; Z_{1:n}) \leq \alpha\} = \alpha
    \end{split}
\end{equation}
when we assume weighted exchangeability according to,
\begin{equation}
    Y_{n+1}(1) - Y_{n+1}(0)|E_{q_{Y(1)} - q_{Y(0)}} \sim \sum_{i=1}^{n_0} \sum_{j=1}^{n_1}  p^{w_0}_i p^{w_j}_i \delta_{q_{Y(1),j} - q_{Y(0),i}} + [1 - \{1-p_{n+1}^{w_1}\}\{1-p_{n+1}^{w_0}\}] \delta_{\infty}
\end{equation}

where $E_{ITE}$ denotes the event that $\{\Psi_1,\ldots,\Psi_{n0*n1+1}\} = \{\psi_1, \ldots, \psi_{n0*n1+1}\}$, $\Psi_i = q_{Y(1), i \mathbin{//} n1} - q_{Y(0), i \mathbin{\%} n0}$ for $i \in [1,n]$, and $\Psi_{n0*n1+1} = Y_{n+1}(1) - Y_{n+1}(0)=ITE_{n+1}$.
\end{restatable}
\end{theobox}
\end{figure*}
While the CCT-learner comes with a finite sample guarantee, the convolution applied during inference (Figure \ref{fig:illustration}) is impractical in some instances. Therefore, we propose the CMC meta-learner, which performs Monte Carlo (MC) sampling during training (and, optionally, fits an additional model). Specifically, we draw samples from each estimated potential outcome distribution to create MC replicates, then use these replicates to produce ITE samples. For each calibration example, we repeat this procedure multiple times, treating the resulting MC ITEs as pseudo ITEs, and train a CATE meta-learner on these samples to form a calibrated predictive distribution of ITEs. Depending on how we incorporate treatment assignment, we recover the S-, T-, and X-learner variants: the S-learner augments the covariates with treatment, the T-learner fits separate outcome models, and the X-learner additionally refits a model on the MC ITE samples. Note that this reduced computation at inference time comes at a cost as the CMC-learner requires two calibration sets.

\paragraph{Epsilon problem} The "Epsilon Problem" highlights a fundamental challenge in the predictive inference of ITE: the joint distribution of counterfactual noise terms, $\varepsilon(0)$ and $\varepsilon(1)$, is unidentifiable as they are never observed simultaneously. Existing approaches implicitly assume independent noise across treatments, or an identical relation, assumptions that are untestable yet critical for valid predictive distributions. We formalize that the ITE distribution is only identifiable if potential outcome marginals are independent or linked by a specified copula (Proposition \ref{prop:epsilon}). Violations of independence—such as correlated noise due to shared confounders—compromise distributional validity, though conservatively valid prediction intervals remain achievable. Simulations (Figure \ref{fig:results_setupA_B_C_D_nie_epsilon}) empirically demonstrate coverage degradation under noise dependency. While partial identification via bounds offers an alternative, we pragmatically adopt the independent copula assumption (making a fourth assumption), aligning with formal causal models, to enable actionable uncertainty quantification despite its philosophical limitations. In Appendix \ref{ap:epsilon}, we further elaborate on this problem.

\paragraph{Universal consistent CCT- and CMC T-learner}
\label{sec:universal}
The introduction of the fourth assumption allows us to prove the existence of a universal consistent CCT- and CMC T-learner, i.e., the conformal predictive distribution approaches the true data-generating distribution as $n \rightarrow \infty$ under any data generating distribution, which we formalize in Theorem \ref{theo:consistency}.

\begin{theobox}
\begin{restatable}{theorem}{theoConsistency}
\label{theo:consistency}
 Assuming independent (potential) outcome noise distributions $\mathbf{\boldsymbol{\varepsilon(0) \perp\!\!\!\perp \varepsilon(1)}}$, and that the object space $\mathcal{X}$ is standard Borel, a universal consistent CCT- and CMC T-learner exists (Algorithm \ref{alg:fit_cct} and \ref{alg:fit_cmc}). 
\end{restatable}
\end{theobox}
The proof for Theorem \ref{theo:consistency} is given in Appendix \ref{ap:proof-universal}. In short, it uses Theorem 3 of \cite{vovkUniversalPredictiveSystems2022}, stating that a universal consistent CPS exists and that the convolution of two independent weakly converging sequences weakly convergence to the independent joint distribution.

\paragraph{Limitations of the CCT-learner and CMC meta-learners} The CCT-learner’s finite-sample calibration guarantee hinges on untestable assumptions: (1) known propensity scores (typically estimated in practice); and (2) weighted exchangeability between pseudo ITE conformity scores and the true ITE $ITE_{n+1}$ (unverifiable due to unobserved counterfactuals). In randomized controlled trials the propensity is known, whereas in observational studies it must be estimated, in which case the theoretical guarantee no longer strictly holds. Nonetheless, empirical results (Figures \ref{fig:edu-ihdp-dist} and \ref{fig:semi-synthetic-dist-acic2016}) suggest that the methods remain robust, and we note that other works providing coverage guarantees for counterfactual prediction intervals \citep{alaaConformalMetalearnersPredictive2023} rely on the same assumption, and approaches \citep{leiConformalInferenceCounterfactuals2021} that allow an amount of mispecification in the propensity modelling guarantee conservative prediction intervals; however, a translation to predictive distributions would mean partialy unidentified distributions.


\section{Experiments}
\subsection{Experimental setup}
\label{sec:exp-setup}
We evaluate our CCT- and CMC meta-learners through 8 different experiments: 6 using synthetic data (Section \ref{sec:synth-exp}) and 2 using semi-synthetic data (Section \ref{sec:semi-exp}). Both learners utilized the scikit-learn RandomForestRegressor with default hyperparameters. We benchmark our approaches against the state-of-the-art approaches in Table \ref{tab:sota-comparison}, using the hyperparameters as specified in these works. Benchmark approaches, such as DiffPO or GANITE, are occasionally omitted from the evaluation plots as their poor performance messes up the clarity of the results. 
The CMC meta-learners always employed 100 MC samples. Given the flexibility of the FCCN and NOFLITE approaches to fit complex noise distribution (e.g., on the semi-synthetic EDU experiment), we implement the CCT-learner with both approaches as base-learners.

\subsection{Results}
\label{sec:exp-results}
We discuss the performance of the distribution estimations of the ITE and potential outcomes, focusing on the CMC T-learner here. Results for the S- and X-learners can be found in Appendix \ref{ap:exp_results} as they are similar to the CMC T-learner, together with the empirical results of the prediction intervals.
\begin{figure}[htbp]
    \centering
    \includegraphics[width=0.85\linewidth]{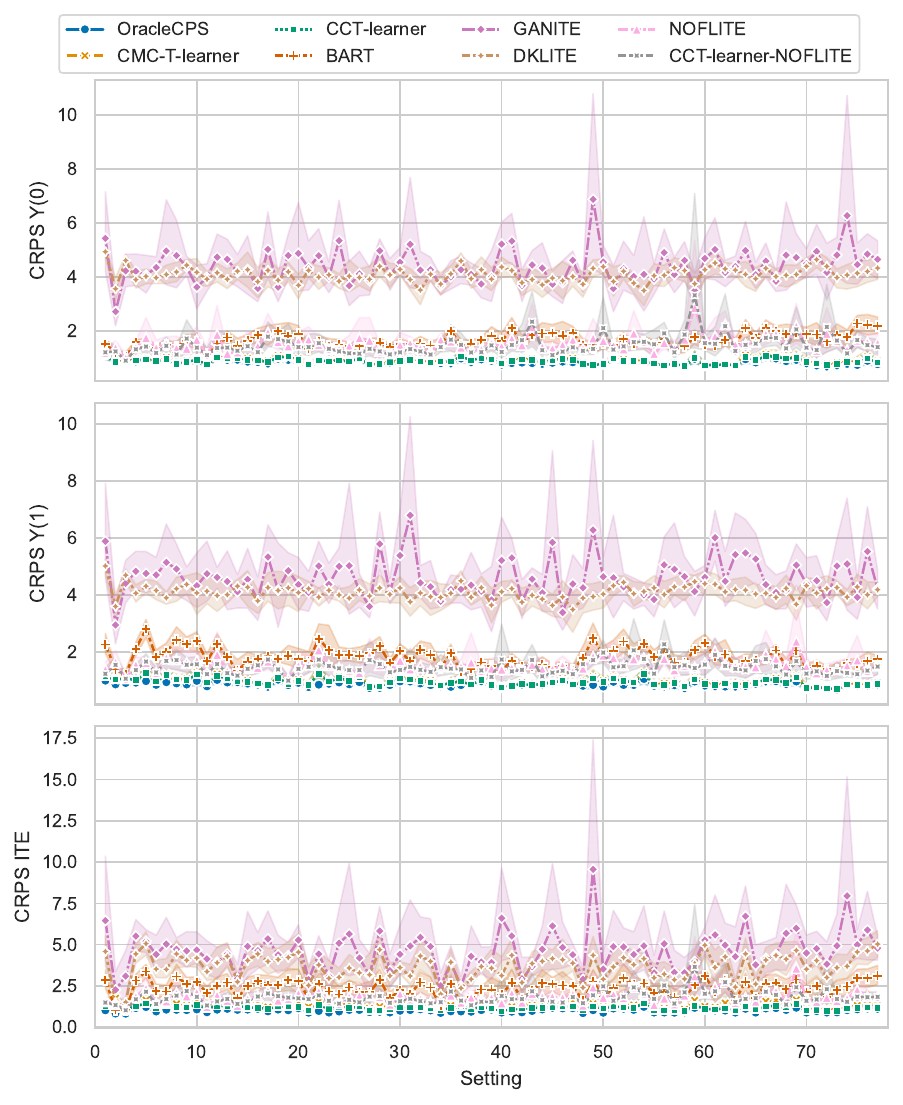}
    \caption{Simulation results for semi-synthetic experiments EDU and ACIC 2016.}
    \label{fig:semi-synthetic-dist-acic2016}
\end{figure}


\paragraph{Probabilistic calibration of CCT- and CMC T-learner}
A predictive distribution is considered probabilistically calibrated if it produces uniformly distributed PIT values on the test set. We assessed the probabilistic calibration of the predictive distribution for all experiments by measuring on the test set the dispersion of the PIT values, i.e., the variance of the PIT values, which should be $\frac{1}{12}$ as this is the dispersion of the standard uniform distribution. Besides measuring the dispersion of the predictive distributions, we also inspect the probabilistic calibration visually for the synthetic experiments from \citet{alaaConformalMetalearnersPredictive2023} and the non-normally distributed semi-synthetic EDU experiment, by plotting the observed frequency of PIT values (Figure \ref{fig:pit-values}). In general, we observe that potential outcome distributions of CCT- and CMC T-learners,  constructed using WCPS, are perfectly probabilistically calibrated given the ground-truth propensity (Figures \ref{fig:alaa-dist}, \ref{fig:nw-A-dist}, \ref{fig:nw-B-dist}, \ref{fig:nw-C-dist}, \ref{fig:nw-D-dist}). 

In our semi-synthetic experiments, we always estimate our propensity scores using logistic regression to represent a more realistic setting. Results show that we can generate probabilistically calibrated predictive distributions even when the propensity score is estimated by checking the dispersion and distribution of PIT-values (Figure \ref{fig:semi-synthetic-dist-acic2016}, \ref{fig:edu-ihdp-dist}, \ref{fig:edu-ihdp-dist}, \ref{fig:pit-values}).

In general, we see that most related work approaches have difficulty generating probabilistically calibrated distributions (Figure \ref{fig:pit-values}). We see that BART and CMGP are reasonably well calibrated when the noise distribution of the experiment is homoscedastic and normally distributed (Figure \ref{fig:pit-values}, Alaa experiments). However, once the noise becomes heteroscedastic and certainly nonnormal, their calibration deteriorates (Figure \ref{fig:edu-ihdp-dist}, \ref{fig:pit-values}). We also observe that the CCT-learner can probabilistically calibrate the FCCN and NOFLITE approaches, often generating slightly overconfident predictive distributions.


\paragraph{CRPS performance} The continuous ranked probability score (CRPS) measures the discrepancy between the predicted CDF and the true outcome, rewarding both calibration (statistical compatibility of probabilistic forecasts and observations) and sharpness (concentration of the distribution around the true value) \citep{matheson1976scoring}. Mathematically defined as $\text{CRPS}(F, y) = \int (F(x) - \mathbbm{1}[x \geq y])^2 dx$, it is a proper scoring rule as formalized by \citet{gneiting2007strictly}. Lower CRPS values indicate better performance.

\begin{table}[htbp]
\centering
\caption{Mean CRPS ranks across potential outcomes and ITE (lower is better)}
\label{tab:crps-ranks}
\resizebox{\linewidth}{!}{
\begin{tabular}{lccc|ccc|ccc|c}
\toprule
\multirow{2}{*}{\textbf{Method}} & \multicolumn{3}{c|}{$ITE$} & \multicolumn{3}{c|}{\textbf{$Y(0)$}} & \multicolumn{3}{c|}{\textbf{$Y(1)$}} & \textbf{Avg. Rank} \\
& Alaa & EDU & ACIC & Alaa & EDU & ACIC & Alaa & EDU & ACIC & \\
\midrule
\textit{\textbf{CCT-learner}  }      & \textbf{1} & 5 & \textbf{1} & 2 & 6 & \textbf{1} & \textbf{1} & 5 & \textbf{1} & \textbf{3.07} \\
\textit{\textbf{CMC-T-learner}  }        & 2 & 6 & 2 & 3 & 7 & 2 & 2 & 7 & 2 & 3.93 \\
BART                   & 5 & 7 & 6 & \textbf{1} & 8 & 5 & 3 & 8 & 5 & 5.00 \\
CEVAE                  & 8 & 10 & 7 & 8 & 9 & 7 & 9 & 9 & 7 & 6.93 \\
CMGP                   & 9 & 9 & 8 & 10 & 11 & 9 & 10 & 11 & 9 & 9.27 \\
DKLITE                 & 11 & 11 & 9 & 11 & 10 & 8 & 11 & 10 & 8 & 9.87 \\
GANITE                 & 10 & 8 & 10 & 9 & 5 & 10 & 6 & 6 & 10 & 8.07 \\
NOFLITE                & 6 & \textbf{1} & 4 & 6 & \textbf{1} & 4 & 7 & \textbf{1} & 4 & 4.13 \\
\textit{\textbf{CCT-learner-NOFLITE}}    & 7 & 2 & 3 & 7 & 2 & 3 & 8 & 2 & 3 & 5.47 \\
DiffPO                 & 12 & 12 & 12 & 12 & 12 & 12 & 12 & 12 & 12 & 12.00 \\
FCCN                   & 3 & 4 & 11 & 5 & 4 & 11 & 5 & 4 & 11 & 6.53 \\
\textit{\textbf{CCT-learner-FCCN} }      & 4 & 3 & 5 & 4 & 3 & 6 & 4 & 3 & 6 & 3.73 \\
\bottomrule
\end{tabular}
}
\end{table}

As summarized in Table~\ref{tab:crps-ranks}, the CCT- and CMC-based learners consistently rank among the top methods across ITEs and counterfactual outcomes. For the ACIC 2016 experiments, the CCT- and CMC T-learners achieve significantly lower CRPS than benchmark methods for potential outcomes and ITE distributions, demonstrating superior accuracy in capturing complex data-generating processes (Figure \ref{fig:semi-synthetic-dist-acic2016}). In the synthetic experiments from \citet{alaaConformalMetalearnersPredictive2023}, setup A (where treatment effects are absent), BART and CMGP perform competitively due to their S-learner structure (modeling treatment as a covariate) and alignment with homoscedastic Gaussian noise assumptions (Figure \ref{fig:alaa-dist}). However, their performance deteriorates in the semi-synthetic EDU experiment where outcomes follow an exponential noise distribution, violating their parametric assumptions (Figure \ref{fig:edu-ihdp-dist}). Here, FCCN and NOFLITE excel by flexibly modeling non-normal noise, and their integration with the CCT-learner further enhances performance by enforcing probabilistic calibration, resulting in even lower CRPS. Notably, FCCN and NOFLITE underperform in certain ACIC 2016 and synthetic settings, likely due to limited training data or suboptimal hyperparameter tuning (Figure \ref{fig:semi-synthetic-dist-acic2016}). Across all experiments, the CCT- and CMC meta-learners consistently strike a balance between calibration and precision, achieving robust CRPS improvements over state-of-the-art alternatives.

\section{Conclusion}
This work presents the CCT- and CMC meta-learners for estimating the predictive distribution of potential outcomes and ITE, enhancing decision-making confidence. To our knowledge, this is the first framework to provide finite-sample guarantees for calibrating predictive distributions in this context. It outperforms related work by demonstrating relatively strong and consistent CRPS scores while achieving probabilistic calibration. Future work could investigate other forms of finite-sample calibration guarantees and the use of our introduced learners in automatic decision-making.

\section*{Acknowledgement}
Part of this research was supported through the Flemish Government (AI Research Program). Jef Jonkers is funded by the Research Foundation Flanders (FWO, Ref.~1S11525N). Jarne Verhaeghe is funded by the Research Foundation Flanders (FWO, Ref.~1S59522N).

\bibliographystyle{unsrtnat}
\bibliography{references}

\begin{thebibliography}{49}
\providecommand{\natexlab}[1]{#1}
\providecommand{\url}[1]{\texttt{#1}}
\expandafter\ifx\csname urlstyle\endcsname\relax
  \providecommand{\doi}[1]{doi: #1}\else
  \providecommand{\doi}{doi: \begingroup \urlstyle{rm}\Url}\fi

\bibitem[Wager and Athey(2018)]{wagerEstimationInferenceHeterogeneous2018}
Stefan Wager and Susan Athey.
\newblock Estimation and {Inference} of {Heterogeneous} {Treatment} {Effects} using {Random} {Forests}.
\newblock \emph{Journal of the American Statistical Association}, 113\penalty0 (523):\penalty0 1228--1242, July 2018.
\newblock \doi{10.1080/01621459.2017.1319839}.

\bibitem[Künzel et~al.(2019)Künzel, Sekhon, Bickel, and Yu]{kunzelMetalearnersEstimatingHeterogeneous2019}
Sören~R. Künzel, Jasjeet~S. Sekhon, Peter~J. Bickel, and Bin Yu.
\newblock Metalearners for estimating heterogeneous treatment effects using machine learning.
\newblock \emph{Proceedings of the National Academy of Sciences}, 116\penalty0 (10):\penalty0 4156--4165, March 2019.
\newblock \doi{10.1073/pnas.1804597116}.

\bibitem[Nie and Wager(2021)]{nieQuasioracleEstimationHeterogeneous2021}
X~Nie and S~Wager.
\newblock Quasi-oracle estimation of heterogeneous treatment effects.
\newblock \emph{Biometrika}, 108\penalty0 (2):\penalty0 299--319, June 2021.
\newblock \doi{10.1093/biomet/asaa076}.

\bibitem[Kennedy(2022)]{kennedyOptimalDoublyRobust2022}
Edward~H. Kennedy.
\newblock Towards optimal doubly robust estimation of heterogeneous causal effects, May 2022.

\bibitem[Curth and Schaar(2021)]{curthNonparametricEstimationHeterogeneous2021}
Alicia Curth and Mihaela van~der Schaar.
\newblock Nonparametric {Estimation} of {Heterogeneous} {Treatment} {Effects}: {From} {Theory} to {Learning} {Algorithms}.
\newblock In \emph{Proceedings of {The} 24th {International} {Conference} on {Artificial} {Intelligence} and {Statistics}}, pages 1810--1818. PMLR, March 2021.

\bibitem[Banerji et~al.(2023)Banerji, Chakraborti, Harbron, and MacArthur]{banerjiClinicalAITools2023}
Christopher R.~S. Banerji, Tapabrata Chakraborti, Chris Harbron, and Ben~D. MacArthur.
\newblock Clinical {AI} tools must convey predictive uncertainty for each individual patient.
\newblock \emph{Nature Medicine}, pages 1--3, October 2023.
\newblock \doi{10.1038/s41591-023-02562-7}.

\bibitem[Feuerriegel et~al.(2024)Feuerriegel, Frauen, Melnychuk, Schweisthal, Hess, Curth, Bauer, Kilbertus, Kohane, and van~der Schaar]{feuerriegel_causal_2024}
Stefan Feuerriegel, Dennis Frauen, Valentyn Melnychuk, Jonas Schweisthal, Konstantin Hess, Alicia Curth, Stefan Bauer, Niki Kilbertus, Isaac~S. Kohane, and Mihaela van~der Schaar.
\newblock Causal machine learning for predicting treatment outcomes.
\newblock \emph{Nature Medicine}, 30\penalty0 (4):\penalty0 958--968, April 2024.
\newblock ISSN 1546-170X.
\newblock \doi{10.1038/s41591-024-02902-1}.
\newblock URL \url{https://www.nature.com/articles/s41591-024-02902-1}.
\newblock Publisher: Nature Publishing Group.

\bibitem[Chipman et~al.(2010)Chipman, George, and McCulloch]{chipman_bart_2010}
Hugh~A. Chipman, Edward~I. George, and Robert~E. McCulloch.
\newblock {BART}: {Bayesian} additive regression trees.
\newblock \emph{The Annals of Applied Statistics}, 4\penalty0 (1):\penalty0 266--298, March 2010.
\newblock ISSN 1932-6157, 1941-7330.
\newblock \doi{10.1214/09-AOAS285}.
\newblock URL \url{https://projecteuclid.org/journals/annals-of-applied-statistics/volume-4/issue-1/BART-Bayesian-additive-regression-trees/10.1214/09-AOAS285.full}.
\newblock Publisher: Institute of Mathematical Statistics.

\bibitem[Hill(2011)]{hillBayesianNonparametricModeling2011a}
Jennifer~L. Hill.
\newblock Bayesian {Nonparametric} {Modeling} for {Causal} {Inference}.
\newblock \emph{Journal of Computational and Graphical Statistics}, 20\penalty0 (1):\penalty0 217--240, January 2011.
\newblock \doi{10.1198/jcgs.2010.08162}.

\bibitem[Alaa and van~der Schaar(2017)]{alaaBayesianInferenceIndividualized2017}
Ahmed~M. Alaa and Mihaela van~der Schaar.
\newblock Bayesian {Inference} of {Individualized} {Treatment} {Effects} using {Multi}-task {Gaussian} {Processes}.
\newblock In \emph{Advances in {Neural} {Information} {Processing} {Systems}}, volume~30, 2017.

\bibitem[Zhang et~al.(2020)Zhang, Bellot, and Schaar]{zhang_learning_2020}
Yao Zhang, Alexis Bellot, and Mihaela Schaar.
\newblock Learning {Overlapping} {Representations} for the {Estimation} of {Individualized} {Treatment} {Effects}.
\newblock In \emph{Proceedings of the {Twenty} {Third} {International} {Conference} on {Artificial} {Intelligence} and {Statistics}}, pages 1005--1014. PMLR, June 2020.
\newblock URL \url{https://proceedings.mlr.press/v108/zhang20c.html}.
\newblock ISSN: 2640-3498.

\bibitem[Louizos et~al.(2017)Louizos, Shalit, Mooij, Sontag, Zemel, and Welling]{louizos_causal_2017}
Christos Louizos, Uri Shalit, Joris~M Mooij, David Sontag, Richard Zemel, and Max Welling.
\newblock Causal effect inference with deep latent-variable models.
\newblock In I.~Guyon, U.~Von Luxburg, S.~Bengio, H.~Wallach, R.~Fergus, S.~Vishwanathan, and R.~Garnett, editors, \emph{Advances in Neural Information Processing Systems}, volume~30. Curran Associates, Inc., 2017.
\newblock URL \url{https://proceedings.neurips.cc/paper_files/paper/2017/file/94b5bde6de888ddf9cde6748ad2523d1-Paper.pdf}.

\bibitem[Yoon et~al.(2018)Yoon, Jordon, and Schaar]{yoon_ganite_2018}
Jinsung Yoon, James Jordon, and Mihaela van~der Schaar.
\newblock {GANITE}: {Estimation} of {Individualized} {Treatment} {Effects} using {Generative} {Adversarial} {Nets}.
\newblock February 2018.
\newblock URL \url{https://openreview.net/forum?id=ByKWUeWA-}.

\bibitem[Vanderschueren et~al.(2023)Vanderschueren, Berrevoets, and Verbeke]{vanderschueren_noflite_2023}
Toon Vanderschueren, Jeroen Berrevoets, and Wouter Verbeke.
\newblock {NOFLITE}: {Learning} to {Predict} {Individual} {Treatment} {Effect} {Distributions}.
\newblock \emph{Transactions on Machine Learning Research}, July 2023.
\newblock ISSN 2835-8856.
\newblock URL \url{https://openreview.net/forum?id=EjqopDxLbG}.

\bibitem[Ma et~al.(2024)Ma, Melnychuk, Schweisthal, and Feuerriegel]{ma_diffpo_2024}
Yuchen Ma, Valentyn Melnychuk, Jonas Schweisthal, and Stefan Feuerriegel.
\newblock {DiffPO}: {A} causal diffusion model for learning distributions of potential outcomes.
\newblock November 2024.
\newblock URL \url{https://openreview.net/forum?id=merJ77Jipt}.

\bibitem[Zhou et~al.(2022)Zhou, Iv, and Carlson]{zhou_estimating_2022}
Tianhui Zhou, William E.~Carson Iv, and David Carlson.
\newblock Estimating {Potential} {Outcome} {Distributions} with {Collaborating} {Causal} {Networks}.
\newblock \emph{Transactions on Machine Learning Research}, June 2022.
\newblock ISSN 2835-8856.
\newblock URL \url{https://openreview.net/forum?id=q1Fey9feu7}.

\bibitem[Lei and Candès(2021)]{leiConformalInferenceCounterfactuals2021}
Lihua Lei and Emmanuel~J. Candès.
\newblock Conformal {Inference} of {Counterfactuals} and {Individual} {Treatment} {Effects}.
\newblock \emph{Journal of the Royal Statistical Society Series B: Statistical Methodology}, 83\penalty0 (5):\penalty0 911--938, November 2021.
\newblock \doi{10.1111/rssb.12445}.

\bibitem[Alaa et~al.(2023)Alaa, Ahmad, and van~der Laan]{alaaConformalMetalearnersPredictive2023}
Ahmed Alaa, Zaid Ahmad, and Mark van~der Laan.
\newblock Conformal {Meta}-learners for {Predictive} {Inference} of {Individual} {Treatment} {Effects}, August 2023.

\bibitem[Vovk et~al.(2019)Vovk, Shen, Manokhin, and Xie]{vovkNonparametricPredictiveDistributions2019}
Vladimir Vovk, Jieli Shen, Valery Manokhin, and Min-Ge Xie.
\newblock Nonparametric predictive distributions based on conformal prediction.
\newblock \emph{Machine Language}, 108\penalty0 (3):\penalty0 445--474, March 2019.
\newblock \doi{10.1007/s10994-018-5755-8}.

\bibitem[Jonkers et~al.(2024)Jonkers, Van~Wallendael, Duchateau, and Van~Hoecke]{jonkers2024conformal}
Jef Jonkers, Glenn Van~Wallendael, Luc Duchateau, and Sofie Van~Hoecke.
\newblock Conformal predictive systems under covariate shift.
\newblock In Simone Vantini, Matteo Fontana, Aldo Solari, Henrik Boström, and Lars Carlsson, editors, \emph{Proceedings of the Thirteenth Symposium on Conformal and Probabilistic Prediction with Applications}, volume 230 of \emph{Proceedings of Machine Learning Research}, pages 406--423. PMLR, 09--11 Sep 2024.
\newblock URL \url{https://proceedings.mlr.press/v230/jonkers24a.html}.

\bibitem[Vovk et~al.(2022)Vovk, Gammerman, and Shafer]{vovkAlgorithmicLearningRandom2022}
Vladimir Vovk, Alexander Gammerman, and Glenn Shafer.
\newblock \emph{Algorithmic {Learning} in a {Random} {World}}.
\newblock Springer International Publishing, Cham, 2022.
\newblock \doi{10.1007/978-3-031-06649-8}.

\bibitem[Tibshirani et~al.(2019)Tibshirani, Foygel~Barber, Candes, and Ramdas]{tibshiraniConformalPredictionCovariate2019a}
Ryan~J Tibshirani, Rina Foygel~Barber, Emmanuel Candes, and Aaditya Ramdas.
\newblock Conformal {Prediction} {Under} {Covariate} {Shift}.
\newblock In \emph{Advances in {Neural} {Information} {Processing} {Systems}}, volume~32, 2019.

\bibitem[Splawa-Neyman et~al.(1990)Splawa-Neyman, Dabrowska, and Speed]{neyman_application_1990}
Jerzy Splawa-Neyman, D.~M. Dabrowska, and T.~P. Speed.
\newblock On the application of probability theory to agricultural experiments. essay on principles. section 9.
\newblock \emph{Statistical Science}, 5\penalty0 (4):\penalty0 465--472, 1990.
\newblock ISSN 08834237, 21688745.
\newblock URL \url{http://www.jstor.org/stable/2245382}.

\bibitem[Rubin(2005)]{rubinCausalInferenceUsing2005}
Donald~B Rubin.
\newblock Causal {Inference} {Using} {Potential} {Outcomes}.
\newblock \emph{Journal of the American Statistical Association}, 100\penalty0 (469):\penalty0 322--331, March 2005.
\newblock \doi{10.1198/016214504000001880}.

\bibitem[Vovk et~al.(2020)Vovk, Petej, Nouretdinov, Manokhin, and Gammerman]{vovkComputationallyEfficientVersions2020}
Vladimir Vovk, Ivan Petej, Ilia Nouretdinov, Valery Manokhin, and Alexander Gammerman.
\newblock Computationally efficient versions of conformal predictive distributions.
\newblock \emph{Neurocomputing}, 397:\penalty0 292--308, July 2020.
\newblock \doi{10.1016/j.neucom.2019.10.110}.

\bibitem[Chernozhukov et~al.(2021)Chernozhukov, Wüthrich, and Zhu]{chernozhukov_distributional_2018}
Victor Chernozhukov, Kaspar Wüthrich, and Yinchu Zhu.
\newblock Distributional conformal prediction.
\newblock \emph{Proceedings of the National Academy of Sciences}, 118\penalty0 (48):\penalty0 e2107794118, 2021.
\newblock \doi{10.1073/pnas.2107794118}.
\newblock URL \url{https://www.pnas.org/doi/abs/10.1073/pnas.2107794118}.

\bibitem[Imbens and Rubin(2015)]{imbens_causal_2015}
Guido~W. Imbens and Donald~B. Rubin.
\newblock \emph{Causal {Inference} for {Statistics}, {Social}, and {Biomedical} {Sciences}: {An} {Introduction}}.
\newblock Cambridge University Press, Cambridge, 2015.

\bibitem[Vovk(2022)]{vovkUniversalPredictiveSystems2022}
Vladimir Vovk.
\newblock Universal predictive systems.
\newblock \emph{Pattern Recognition}, 126:\penalty0 108536, June 2022.
\newblock ISSN 0031-3203.
\newblock \doi{10.1016/j.patcog.2022.108536}.

\bibitem[Matheson and Winkler(1976)]{matheson1976scoring}
James~E Matheson and Robert~L Winkler.
\newblock Scoring rules for continuous probability distributions.
\newblock \emph{Management science}, 22\penalty0 (10):\penalty0 1087--1096, 1976.

\bibitem[Gneiting and Raftery(2007)]{gneiting2007strictly}
Tilmann Gneiting and Adrian~E Raftery.
\newblock Strictly proper scoring rules, prediction, and estimation.
\newblock \emph{Journal of the American statistical Association}, 102\penalty0 (477):\penalty0 359--378, 2007.

\bibitem[Papadopoulos et~al.(2002)Papadopoulos, Proedrou, Vovk, and Gammerman]{papadopoulosInductiveConfidenceMachines2002}
Harris Papadopoulos, Kostas Proedrou, Volodya Vovk, and Alex Gammerman.
\newblock Inductive {Confidence} {Machines} for {Regression}.
\newblock In Tapio Elomaa, Heikki Mannila, and Hannu Toivonen, editors, \emph{Machine {Learning}: {ECML} 2002}, Lecture {Notes} in {Computer} {Science}, pages 345--356, Berlin, Heidelberg, 2002. Springer.
\newblock \doi{10.1007/3-540-36755-1_29}.

\bibitem[Vovk et~al.(2003)Vovk, Lindsay, Nouretdinov, and Gammerman]{vovkMondrianConfidenceMachine2003}
Vladimir Vovk, David Lindsay, Ilia Nouretdinov, and Alex Gammerman.
\newblock Mondrian {Confidence} {Machine}, March 2003.

\bibitem[Romano et~al.(2019)Romano, Patterson, and Candes]{romanoConformalizedQuantileRegression2019}
Yaniv Romano, Evan Patterson, and Emmanuel Candes.
\newblock Conformalized {Quantile} {Regression}.
\newblock In \emph{Advances in {Neural} {Information} {Processing} {Systems}}, volume~32. Curran Associates, Inc., 2019.

\bibitem[Papadopoulos and Haralambous(2011)]{papadopoulosReliablePredictionIntervals2011}
Harris Papadopoulos and Haris Haralambous.
\newblock Reliable prediction intervals with regression neural networks.
\newblock \emph{Neural Networks}, 24\penalty0 (8):\penalty0 842--851, October 2011.
\newblock \doi{10.1016/j.neunet.2011.05.008}.

\bibitem[Angelopoulos and Bates(2022)]{angelopoulosGentleIntroductionConformal2022}
Anastasios~N. Angelopoulos and Stephen Bates.
\newblock A {Gentle} {Introduction} to {Conformal} {Prediction} and {Distribution}-{Free} {Uncertainty} {Quantification}.
\newblock Technical Report arXiv:2107.07511, arXiv, January 2022.

\bibitem[Fontana et~al.(2022)Fontana, Zeni, and Vantini]{fontanaConformalPredictionUnified2022}
Matteo Fontana, Gianluca Zeni, and Simone Vantini.
\newblock Conformal {Prediction}: a {Unified} {Review} of {Theory} and {New} {Challenges}, July 2022.

\bibitem[Manokhin(2022)]{manokhin_valery_2022_6467205}
Valery Manokhin.
\newblock Awesome conformal prediction, April 2022.

\bibitem[Angelopoulos et~al.(2024)Angelopoulos, Barber, and Bates]{angelopoulos2024theoreticalfoundationsconformalprediction}
Anastasios~N. Angelopoulos, Rina~Foygel Barber, and Stephen Bates.
\newblock Theoretical foundations of conformal prediction, 2024.
\newblock URL \url{https://arxiv.org/abs/2411.11824}.

\bibitem[Rissanen and Marttinen(2021)]{rissanen2021a}
Severi Rissanen and Pekka Marttinen.
\newblock A critical look at the consistency of causal estimation with deep latent variable models.
\newblock In A.~Beygelzimer, Y.~Dauphin, P.~Liang, and J.~Wortman Vaughan, editors, \emph{Advances in Neural Information Processing Systems}, 2021.
\newblock URL \url{https://openreview.net/forum?id=vU96vWPrWL}.

\bibitem[Zhou et~al.(2021)Zhou, Li, Wu, and Carlson]{zhou_estimating_2021}
Tianhui Zhou, Yitong Li, Yuan Wu, and David Carlson.
\newblock Estimating {Uncertainty} {Intervals} from {Collaborating} {Networks}.
\newblock \emph{Journal of Machine Learning Research}, 22\penalty0 (257):\penalty0 1--47, 2021.
\newblock ISSN 1533-7928.
\newblock URL \url{http://jmlr.org/papers/v22/20-1100.html}.

\bibitem[Sklar(1959)]{sklar1959fonctions}
M~Sklar.
\newblock Fonctions de r{\'e}partition {\`a} n dimensions et leurs marges.
\newblock In \emph{Annales de l'ISUP}, volume~8, pages 229--231, 1959.

\bibitem[Robins(1986)]{ROBINS19861393}
James Robins.
\newblock A new approach to causal inference in mortality studies with a sustained exposure period---application to control of the healthy worker survivor effect.
\newblock \emph{Mathematical Modelling}, 7\penalty0 (9):\penalty0 1393--1512, 1986.
\newblock ISSN 0270-0255.
\newblock \doi{https://doi.org/10.1016/0270-0255(86)90088-6}.
\newblock URL \url{https://www.sciencedirect.com/science/article/pii/0270025586900886}.

\bibitem[Melnychuk et~al.(2024)Melnychuk, Feuerriegel, and van~der Schaar]{melnychuk2024quantifying}
Valentyn Melnychuk, Stefan Feuerriegel, and Mihaela van~der Schaar.
\newblock Quantifying aleatoric uncertainty of the treatment effect: A novel orthogonal learner.
\newblock In \emph{The Thirty-eighth Annual Conference on Neural Information Processing Systems}, 2024.
\newblock URL \url{https://openreview.net/forum?id=RDsDvSHGkA}.

\bibitem[McLeish(2005)]{mcleishSTAT901Probability2005}
Don McLeish.
\newblock \emph{{{STAT}} 901: {{Probability}}}.
\newblock University of Waterloo, 2005.
\newblock URL \url{https://sas.uwaterloo.ca/~dlmcleis/s901/s901_2005.pdf}.

\bibitem[Dorie et~al.(2019)Dorie, Hill, Shalit, Scott, and Cervone]{dorie_automated_2019}
Vincent Dorie, Jennifer Hill, Uri Shalit, Marc Scott, and Dan Cervone.
\newblock Automated versus {Do}-{It}-{Yourself} {Methods} for {Causal} {Inference}: {Lessons} {Learned} from a {Data} {Analysis} {Competition}.
\newblock \emph{Statistical Science}, 34\penalty0 (1):\penalty0 43--68, February 2019.

\bibitem[Carvalho et~al.(2019)Carvalho, Feller, Murray, Woody, and Yeager]{carvalhoAssessingTreatmentEffect2019}
Carlos Carvalho, Avi Feller, Jared Murray, Spencer Woody, and David Yeager.
\newblock Assessing {Treatment} {Effect} {Variation} in {Observational} {Studies}: {Results} from a {Data} {Challenge}.
\newblock \emph{Observational Studies}, 5\penalty0 (2):\penalty0 21--35, 2019.

\bibitem[Shalit et~al.(2017)Shalit, Johansson, and Sontag]{shalitEstimatingIndividualTreatment2017}
Uri Shalit, Fredrik~D. Johansson, and David Sontag.
\newblock Estimating individual treatment effect: generalization bounds and algorithms.
\newblock In \emph{Proceedings of the 34th {International} {Conference} on {Machine} {Learning}}, pages 3076--3085. PMLR, July 2017.

\bibitem[Banerji et~al.(2017)Banerji, Berry, and Shotland]{banerji2017impact}
Rukmini Banerji, James Berry, and Marc Shotland.
\newblock The impact of maternal literacy and participation programs: Evidence from a randomized evaluation in india.
\newblock \emph{American Economic Journal: Applied Economics}, 9\penalty0 (4):\penalty0 303--337, 2017.

\bibitem[Curth and van~der Schaar(2021)]{curth_inductive_2021}
Alicia Curth and Mihaela van~der Schaar.
\newblock On {Inductive} {Biases} for {Heterogeneous} {Treatment} {Effect} {Estimation}.
\newblock In \emph{Advances in {Neural} {Information} {Processing} {Systems}}, volume~34, pages 15883--15894. Curran Associates, Inc., 2021.

\end{thebibliography}

\appendix

\newpage
\appendix
\onecolumn
\section{Background}
\subsection{Conformal prediction}
\label{ap:cp}
Conformal prediction (CP) is a model-agnostic framework that gives an implicit confidence estimate in a prediction by generating prediction sets at a specified significance level $\alpha$~\citep{vovkAlgorithmicLearningRandom2022}. Notably, this framework allows for (conservatively) valid non-asymptotic confidence predictors under the exchangeability assumption, assuming training/calibration data exchangeability with test data, a slightly less stringent assumption than the IID assumption. The prediction sets in CP are formed by comparing nonconformity scores of examples that quantify how unusual a predicted label is, i.e., these scores measure the disagreement between the prediction and the actual target.\\
More specifically, we define a prediction interval $\hat{C}_\alpha(X_{n+1})$, for test object $X_{n+1} \in \mathcal{X}$, by computing the conformity scores $S_i^y$, based on conformity score $s$ for each $y \in \mathbb{R}$, using training data $Z_i = (X_i, Y_i), i=1,...,n$:
\begin{equation}
    \begin{split}
        \hat{C}_{\alpha}&(X_{n+1}) = \Bigl\{ \\
        &y \in \mathbb{R}: \frac{|i=1,...,n+1: s((X_i, Y_i); Z_{1:n \setminus i} \cup  (X_{n+1}, y)) \geq s((X_{n+1},y);Z_{1:n})|}{n+1} > \alpha \\
        \Bigl\} &
    \end{split}
\end{equation}
This procedure, termed full or transductive CP, is computationally intensive, as it requires refitting of the underlying model for each $y$ in the target domain. Therefore, a more applicable variant of full CP, called inductive or split CP (ICP) \citep{papadopoulosInductiveConfidenceMachines2002}, offers computational efficiency and enables the integration of CP with ML algorithms, such as neural networks and tree-based models. In ICP, the training sequence $Z_{1:n} = \{(X_1, Y_1), ...,(X_{n}, Y_{n})\}$ is portioned into a proper training sequence $\{(X_1, Y_1), ..., (X_{m}, Y_{m})\}$ and a calibration sequence $\{(X_{m+1}, Y_{m+1}), ..., (X_{n}, Y_{n})\}$. We first train a regression model $\hat{y}(x)$ using the proper training sequence. Then, based on nonconformity score $s$, we generate nonconformity scores $S_i$ for data points $(X_i, Y_i)$  where  $m < i \leq n$. The nonconformity score function $s$ leverages the pre-trained regression model, e.g., the absolute error $s=|y-\hat{y}|$. These nonconformity scores together with test object $X_{n+1}$ and significance level $\alpha$ defines:
\begin{equation}
        \hat{C}_{\alpha}(X_{n+1}) = \left\{ y \in \mathbb{R}: \frac{|i=m+1,...,n+1|: s(X_{i}, Y_{i}) \geq s(X_{n+1},y)|}{n+1} > \alpha \right\}
\end{equation}
Although the conventional ICP yields calibrated prediction intervals, it only offers marginal coverage and constant interval sizes for different covariates. As we want more conditional valid prediction intervals or at least smaller intervals for confident predictions and larger intervals for more "difficult" examples, extensions of ICP are proposed, such as Mondrian CP~\citep{vovkMondrianConfidenceMachine2003} by adjusting the CP algorithm and conformalized quantile regression (CQR)~\citep{romanoConformalizedQuantileRegression2019} by introducing a clever nonconformity score. Another adaptation of ICP for establishing more adaptive prediction intervals involves employing normalized nonconformity score~\citep{papadopoulosReliablePredictionIntervals2011}. These normalized scores typically use the product of the absolute error $|y-\hat{f}(x)|$ and the reciprocal of a 1D uncertainty estimate $\frac{1}{\hat{u}(x)}$, expressed as $s = \frac{|y-\hat{f}(x)|}{\hat{u}(x)}$, where $\hat{u}(x)$ may be generated by a distinct model which estimates $|y-\hat{f}(x)|$. For a more comprehensive exploration of CP, its extensions and applications, we refer the reader to other works~\citep{vovkAlgorithmicLearningRandom2022, angelopoulosGentleIntroductionConformal2022, fontanaConformalPredictionUnified2022, manokhin_valery_2022_6467205, angelopoulos2024theoreticalfoundationsconformalprediction}.



\newpage
\section{Related work}
\subsection{CATE meta-learners}
Conditional average treatment effect (CATE) estimation is frequently performed using CATE meta-learners. CATE meta-learners are frameworks where base learners are utilized to learn and perform CATE estimation~\citep{kunzelMetalearnersEstimatingHeterogeneous2019}.
Some examples are the T-learner, S-learner, X-learner~\citep{kunzelMetalearnersEstimatingHeterogeneous2019}, DR-learner~\citep{kennedyOptimalDoublyRobust2022}, and R-learner~\citep{nieQuasioracleEstimationHeterogeneous2021}. We can categorize these learners into two groups: one-step plug-in (indirect) learners and two-step (direct) learners.  One-step plug-in learners (S- and T-learners) are approaches that use two regression functions where the difference represents the CATE function and thus is estimated indirectly. In contrast, two-step learners (X-, DR-, and R-learners) output the CATE function directly. Three learners are used in this work: the T-, S-, and X-learner. In the binary treatment case, the T-learner fits two different base learners $\mu_0$ and $\mu_1$ on data without treatment and with treatment, respectively. The CATE is then estimated as $\hat{\tau}(X)=\hat{\mu}_1(X)-\hat{\mu}_0(X)$. In contrast, the S-learner fits a single base learner $\mu$ on $[X, W]$ with $W$ the treatment variable. The CATE is then estimated as $\hat{\tau}(X)=\hat{\mu}(X,W=1)-\hat{\mu}(X,W=0)$. The X-learner starts as a T-learner by first fitting $\mu_0$ and $\mu_1$, however, afterwards a direct estimate of $\tau$ is fitted with $X$ as input and $W(Y-\hat{\mu}_0(X)) + (1-W)(\hat{\mu}_1(X) - Y)$ as target.

\subsection{Prediction intervals for ITE}
This section provides a more comprehensive overview of the literature on predictive inference of ITE and quantifying related predictive uncertainty through prediction intervals. Uncertainty quantification is a crucial component for treatment effect estimations for reliable decision-making in high-risk environments, such as healthcare~\citep{banerjiClinicalAITools2023}. In the past, most proposals for uncertainty quantification in ITE estimation used Bayesian approaches, such as Bayesian additive regression trees (BART)~\citep{hillBayesianNonparametricModeling2011a} and Gaussian processes~\citep{alaaBayesianInferenceIndividualized2017}. The problem with these Bayesian approaches is that they are often uncalibrated without knowing the underlying data-generating process. These Bayesian approaches are also model-specific and are not easily transferred to modern ML approaches.

Recently, some works have tried to resolve this issue by using the CP framework ~\citep{leiConformalInferenceCounterfactuals2021, alaaConformalMetalearnersPredictive2023}, which can provide non-asymptotic distribution-free coverage guarantees and thus resolves the significant shortcomings of these Bayesian approaches. \citet{leiConformalInferenceCounterfactuals2021} was the first work to propose a conformal inference approach to produce reliable prediction intervals for ITEs. More specifically, they proposed using WCP to account for covariate shifts as the distributions of covariates for treated and untreated subjects can differ from the target population. They proposed three variants of WCP ITE algorithms: (1) Naive WCP, which combines two potential outcome interval estimates, estimated by applying both WCP and a Bonferroni correction to get an ITE prediction interval; (2) Exact nested WCP, which first generates plug-in estimates of ITEs by using WCP and uses these estimates to apply a regular CP; and (3) Inexact nested WCP which replaces the second CP step in Exact WCP with quantile regression which does not provide any coverage guarantees. A significant drawback of the exact and inexact WCP approaches is that they require refitting the model, at least partially, for different confidence levels for the prediction intervals. Additionally, all the proposed approaches based on WCP only provide intervals as uncertainty quantification.

\citet{alaaConformalMetalearnersPredictive2023} propose conformal meta (CM)-learners, which is a model-agnostic framework for issuing predictive intervals for ITEs that applies the inductive CP on top of CATE meta-learners. These CM-learners are applied to the previously discussed direct meta-learners, which use a two-step procedure where nuisance estimators are fitted on the outcome in the first step. In the second step, these nuisance estimators are used to generate a pseudo outcome in which the covariates are regressed on to get point estimates of the CATE. To construct prediction intervals of the ITE, \citet{alaaConformalMetalearnersPredictive2023} propose to use these pseudo outcomes to generate conformity scores together with the estimates of the meta-learner, i.e., CATE estimator. However, their proposed framework requires more stringent assumptions to be valid than the classical ones for estimating CATE. They developed the so-called stochastic ordering framework and showed that CM-learners are marginally valid if their pseudo outcome conformity scores are stochastically larger than the "oracle" conformity scores evaluated on the unobserved ITEs. However, it is difficult to translate this into practice, as the only way to verify this assumption is to observe the true ITEs, which is impossible. Additionally, their method is limited to producing prediction intervals of ITE only. Note that the same stochastic ordering framework could be integrated with our framework to get validity guarantees of ITE intervals under the same assumption as the CM-learners. 

\subsection{Potential outcome and ITE predictive distributions}
Understanding potential outcomes and ITE predictive distributions is crucial for causal inference, enabling a more nuanced estimation of treatment effects beyond point estimates and prediction intervals. Various methodologies have been proposed to estimate these distributions, leveraging probabilistic models, deep learning techniques, and generative approaches. This section explores existing methods for constructing potential outcomes and ITE predictive distributions, highlighting their underlying assumptions, strengths, and limitations. Table \ref{tab:sota-comparison} presents a comparative summary of these methods.

\subsubsection*{BART}
Bayesian additive regression trees (BART) \citep{chipman_bart_2010} have been widely used for causal inference, as proposed by \citet{hillBayesianNonparametricModeling2011a}. This approach uses a Bayesian semiparametric modeling procedure to model the posterior distributions of potential outcomes under the assumption of normally distributed noise. Although \citet{hillBayesianNonparametricModeling2011a} does not explicitly model the predictive ITE distribution, it can be obtained by sampling from the counterfactual posterior distribution and assuming independence between potential outcome distributions. This approach for constructing the ITE distribution can be leveraged for any approach that allows sampling of the potential outcome distributions.

\subsubsection*{CMGP}
\citet{alaaBayesianInferenceIndividualized2017} introduce causal multi-task Gaussian processes (CMGPs) to provide credible intervals with Bayesian coverage. Their approach models potential outcomes as the outputs of a function in a vector-valued reproducing kernel Hilbert space (vvRKHS). A multi-task Gaussian process (GP), with a linear coregionalization kernel as prior over the vvRKHS, is used for learning potential outcome and ITE predictive distributions. They deal with confounding by using a risk-based empirical Bayes method for adapting the multi-task GP prior that jointly minimizes the empirical error in the observed outcome. While \citet{alaaBayesianInferenceIndividualized2017} does not explicitly model the ITE predictive distribution, the framework can be extended for this purpose, given its reliance on the error function of the normal distribution. However, this extension remains constrained by prior choices and the normality assumption. The ITE predictive distribution in this setting is given by:
\begin{equation}
    \widehat{ITE} \sim \mathcal{N}(\hat{\tau}(X_{n+1}), V(X_{n+1}))
\end{equation}
where $\hat{\tau}(x) = (K_{\theta^*}^{T}(x)\Lambda_{\theta^*}Y_{1:n})^T [-1, 1]^T$ is the CATE estimate, $V(x) = K_{\theta^*}(x,x) - K_{\theta^*}(x)\Lambda_{\theta^*}K_{\theta^*}^T(x)$ represents the predictive variance, $K_{\theta^*}$ the kernel function parameterized by a set of hyperparameters $\theta^*$, and $\Lambda_{\theta^*}$ represents the inverse covariance matrix of the training data $X_{1:n}$, plus noise (aleatoric uncertainty). Note that while they do not discuss the relation between the potential outcome noise distributions, they assume independence in their experiments. However, we believe their approach allows for encoding different assumptions of the relation by encoding it into a covariance matrix as a prior.

\subsubsection*{CEVAE}
Another approach to estimate potential outcome distributions involves the causal effect variational autoencoder (CEVAE) \citep{louizos_causal_2017}, which employs deep latent variable modeling and variational autoencoders to infer causal effects. While this approach was designed to deal with hidden confounders, it can also be used in settings without hidden confounders; however, consistency issues are noted in the case of misspecified latent variables or complex data distributions \citep{rissanen2021a}. This method also assumes a normal predictive distribution for potential outcomes.

\subsubsection*{GANITE}
While designed to generate predictive ITE distributions,  the approach of \citet{yoon_ganite_2018}, termed generative adversarial nets (GAN) for inference of ITE (GANITE), could also generate potential outcome predictive distributions. GANITE first trains a GAN to generate pseudo counterfactual outcomes, which are then used to train a second GAN. This results in a generator $I(x)$ intended to approximate $(Y(0), Y(1))$. A predictive distribution could be derived using kernel density estimation or additional distributional assumptions. However, the joint distribution obtained from this approach is likely unreliable due to GANITE's inability to capture dependencies between counterfactual noise distributions (Section \ref{ap:epsilon}) or implicitly assume a joint distribution of the noise.

\subsubsection*{DKLITE}
Deep kernel learning for ITEs (DKLITE) \citep{zhang_learning_2020} combines deep kernels with posterior regularization to estimate potential outcomes. This method enables the construction of a potential outcome predictive distribution under the normality assumption. While the authors do not discuss the idea of an ITE predictive distribution, this would only be possible by assuming the conditional independence between the counterfactual predictive distributions and applying the convolution operation or approximating it through sampling. 

\subsubsection*{NOFLITE}
NOFLITE \citep{vanderschueren_noflite_2023} employs normalizing flows to optimize the likelihood of potential outcome distributions. This method mitigates confounding by transforming covariates into a balanced representation and explicitly estimates the ITE distribution through Monte Carlo sampling of potential outcome distributions. NOFLITE does not make explicit assumptions about noise distributions but implicitly assumes potential outcome noise independence through its sampling procedure. However, the method focuses on optimizing the likelihood of the conditional prior, capturing aleatoric uncertainty while disregarding epistemic uncertainty.

\subsubsection*{DiffPO}
DiffPO \citep{ma_diffpo_2024} introduces a causal diffusion model for learning potential outcome distributions. The model incorporates an orthogonal diffusion loss to address confounding. However, its evaluation primarily compares against approaches using MC dropout, which only accounts for approximation-based epistemic uncertainty. While DiffPO effectively quantifies aleatoric uncertainty and evaluates the alignment between estimated and true aleatoric uncertainty, this evaluation may not fully reflect the comparative performance of the model. We see in our evaluations that this approach fails in most settings; however, this could also be due to insufficient hyperparameter tuning.

\subsubsection*{FCCN}
Full adjustment collaborating causal networks (FCCN) \citep{zhou_estimating_2022} estimate conditional potential outcome distributions while addressing covariate shift through latent representation learning. Based on the collaborating networks framework \citep{zhou_estimating_2021}, CCN jointly optimizes two neural networks: one for learning cumulative distributions and another for learning their inverses. Under Lipschitz continuity assumptions, the authors prove that the model converges asymptotically to the true outcome distribution.

\newpage
\section{Conformal Monte Carlo meta-learners}
\label{ap:cmc}
We propose two methods for generating MC ITE samples: (1) The MC ITE method independently samples from both conditional potential outcome predictive distributions, subtracting the potential outcome under control from the one under treatment; (2) The pseudo-MC ITE method adapts this by conditioning on the observed treatment value, calculating the ITE sample as the difference between the observed outcome $Y_i$ and the conditional potential outcome predictive distribution under control, and vice versa when treatment is not administered. Since the two approaches sample from different covariate distributions, they require a different weighting to be applied to WCPS. While the MC requires the same weighting scheme as in the CCT-learner, the pseudo-MC procedure requires a weighting of $\frac{\pi(x)}{1-\pi(x)}$ and $\frac{1-\pi(x)}{\pi(x)}$ for the counterfactual outcome of the treated and untreated subjects. Algorithm~\ref{alg:pseudo_MC_ite} provides a comprehensive overview of this sampling scheme.
\begin{table}[H]
\centering
\caption{Formulas to construct MC ITE sample, where W represents the observed treatment action, Y the observed outcome, and $\hat{Y}(0)$, $\hat{Y}(1)$ are the sampled potential outcome from the conditional potential outcome predictive distribution.}
\label{tab:mc-sample}
\begin{tabular}{@{}llll@{}}
\toprule
                       & & $\mathbf{w_0(x)}$ & $\mathbf{w_1(x)}$ \\
\cmidrule{3-4}
\textbf{\textit{MC} }       & $\hat{Y}(1) - \hat{Y}(0)$ & $\frac{1}{1-\pi(x)}$ & $\frac{1}{\pi(x)}$               \\
\textbf{\textit{pseudo-MC}} & $W(Y - \hat{Y}(0)) + (1-W)(\hat{Y}(1) - Y)$ & $\frac{\pi(x)}{1-\pi(x)}$ & $\frac{1-\pi(x)}{\pi(x)}$ \\\bottomrule
\end{tabular}
\end{table}

\begin{algorithm}[H]
   \caption{(Pseudo) Monte Carlo ITE sampling}
   \label{alg:pseudo_MC_ite}
\begin{algorithmic}[1]
   \STATE {\bfseries Input:} observations $\{Z_i = (Y_i, X_i, W_i)\}_{i=1}^n$, predictive distributions $Q_{Y(0)}$ and $Q_{Y(1)}$, MC samples $n_{MC}$
   \STATE $ITE_{MC} = zeros(n \cdot n_{MC})$

   \FOR{$i=1$ {\bfseries to} $n$}
       \IF{Pseudo CMC}
           \IF{$W_i == 0$}
               \STATE Sample $u_1 \sim Uniform(0,1, n_{MC} \text{ times})$
               \STATE $ITE_{MC,i\cdot n_{MC} : (i+1)n_{MC}}$ = $\textbf{Q}^{-1}_{Y(1)}(X_i,u_1, w_1) - Y_i$ \COMMENT{where $Q^{-1}(x, u, w) = \inf\{y: Q(x, y, w) \geq u\}$ and uses weights specified in Table \ref{tab:mc-sample}}
           \ELSE
               \STATE Sample $u_0 \sim Uniform(0,1, n_{MC} \text{ times})$
               \STATE $ITE_{MC,i\cdot n_{MC} : (i+1)n_{MC}}= y_i - \textbf{Q}^{-1}_{Y(0)}(X_i, u_0, w_0)$
           \ENDIF
           
       \ELSE
       
           \STATE Sample $u_0 \sim Uniform(0,1, n_{MC} \text{ times})$
           \STATE Sample $u_1 \sim Uniform(0,1, n_{MC} \text{ times})$
           \STATE $ITE_{MC,i\cdot n_{MC} : (i+1)n_{MC}} = \textbf{Q}^{-1}_{Y(1)}(X_i,u_1,w_1) - \textbf{Q}^{-1}_{Y(0)}(X_i,u_0,w_0)$
       
       \ENDIF
   \ENDFOR
   \STATE \textbf{return} $ITE_{MC}$
\end{algorithmic}
\end{algorithm}

\subsection{S-learner}
The CMC S-learner closely resembles the T-learner algorithm with a notable distinction: it employs an S-learner as a CATE estimator and a single regression model for modeling the nuisance parameters. This approach offers the advantage of leveraging a single model, thereby increasing the availability of samples to learn common components of the outcome. However, a drawback with this method arises from shared epistemic uncertainty, accounted for twice, potentially compromising the validity of the ITE predictive distributions. Nonetheless, the method remains capable of furnishing conservatively valid prediction intervals even in this scenario, as we observe in the simulation experiments.

\subsection{X-learner}
The procedure for the CMC X-learner mirrors that of the CMC-T-learner, with a key distinction: instead of employing the T-learner as the CATE estimator, we incorporate an extra estimator by fitting a regression model on the generated MC ITE samples. A notable advantage is that the ITE regression model, serving as the CATE estimator, can utilize a reduced set of covariates compared to the input data for the nuisance models. This becomes particularly advantageous in scenarios where only a limited covariate set is accessible during inference. However, a potential drawback is the likelihood of sacrificing some level of validity, as the MC ITE samples derived from the nuisance models may introduce epistemic variability. Consequently, this variability might contribute to widening the predictive distribution. Nevertheless, despite this potential trade-off, the approach still enables the generation of conservatively valid prediction intervals, as we observe in the simulation experiments later presented in this work.  

\newpage
\section{The Epsilon Problem: (un)identifiability of ITE predictive distribution}
\label{ap:epsilon}
Beyond introducing the CCT and CMC meta-learners, we identify a fundamental challenge for predictive inference in ITE estimation: the unknown relationship between the noise distributions under treatment, $\varepsilon(1)$, and control, $\varepsilon(0)$. We propose a fourth assumption, in addition to the standard three outlined in Section~\ref{sec:prob-def}, concerning the joint distribution of these noise terms. This assumption is critical for generating valid and calibrated predictive distributions within our framework.

Since $\varepsilon(0)$ and $\varepsilon(1)$ are never jointly observed, their joint distribution is not identifiable from data. Thus, any distributional regression approach requires an assumption about how these noise terms relate. In practice, most CATE literature assumes the noise terms are zero-mean, independent of covariates $(X, W)$, and either independent or identical across treatment arms~\citep{kunzelMetalearnersEstimatingHeterogeneous2019, nieQuasioracleEstimationHeterogeneous2021, kennedyOptimalDoublyRobust2022, curthNonparametricEstimationHeterogeneous2021}.  Simulation studies often further assume either deterministic treatment effects (implying $\varepsilon(0) = \varepsilon(1)$) \citep{nieQuasioracleEstimationHeterogeneous2021, kennedyOptimalDoublyRobust2022, curthNonparametricEstimationHeterogeneous2021} or full independence of the two noise terms \citep{kunzelMetalearnersEstimatingHeterogeneous2019}. The limited literature concerning predictive inference of ITE ~\cite{hillBayesianNonparametricModeling2011a, alaaBayesianInferenceIndividualized2017, leiConformalInferenceCounterfactuals2021, alaaConformalMetalearnersPredictive2023, vanderschueren_noflite_2023} mostly assumes that the noise distributions are fully independent and sets up their experiments in the same way. 

We formally state the identifiability constraint:
\begin{theobox}
\begin{proposition} 
\label{prop:epsilon}
The ITE distribution $F_{Y(1)-Y(0)|X}(y)$ is only identifiable if \begin{equation} F_{Y(0)|X}(y) \perp\!\!\!\perp F_{Y(1)|X}(y) \end{equation} or if a specific copula linking $Y(0)$ and $Y(1)$ is assumed. 
\end{proposition}
\end{theobox}
\begin{proof}
    By definition,
    \begin{equation}
        F_{Y(1)-Y(0)|X}(y) = \int \int \mathbbm{1}[y_1 - y_0 \leq y] dF_{Y(1),Y(0)}(y_1,y_0)
    \end{equation}
    Thus, the CDF of the difference depends on the full joint distribution $F_{Y(1),Y(0)}(y_1,y_0)$, not just on the two marginals.

    Sklar's theorem \citep{sklar1959fonctions} tells us that any joint CDF can be written in terms of its marginals and a copula $C$:
    \begin{equation}
        F_{Y(1),Y(0)|X}(y_1,y_0) = C\left(F_{Y(1)|X}(y_1), F_{Y(0)|X}(y_0)\right).
    \end{equation}
    Hence the difference CDF is 
    \begin{equation}
         F_{Y(1)-Y(0)|X}(y) = \int \int \mathbbm{1}[y_1 - y_0 \leq y] dC(u_1,u_0) \quad \text{where } u_1=F_{Y(1)|X}(y_1), u_0=F_{Y(0)|X}(y_0).
    \end{equation}
    Thus, without specifying $C$, this integral, and hence $F_{Y(1)-Y(0)|X}(y)$, is not determined by the marginals alone.

    Next, we present two cases where the ITE distribution is identifiable and one where it is not. 
    
    \textit{Case 1. Independence copula:} If we assume $Y(0) \perp\!\!\!\perp Y(1) | X$, then $C(u_1,u_0)$, and thus,
    \begin{equation}
        F_{Y(1)-Y(0)|X}(y) = \int \int \mathbbm{1} [y_1 - y_0] dF_{Y(1)|X}(y_1) dF_{Y(0)|X}(y_0)
    \end{equation}
    which reduces to a convolution of two known marginals (where one is negated), and is therefore identifiable.
    
    \textit{Case 2. Know copula:} Generally, if one assumes a particulate copula $C$, then the joint law is specified by the two marginals and $C$, so the difference CDF becomes identifiable via the same integral above.

    \textit{Case 3. No identified copula} If neither the independence nor a known copula is assumed, then infinitely many different copulas are compatible with the same pair of marginals but induce different distributions of the difference $Y(1) - Y(0)$.

    Therefore, no unique $F_{Y(1) - Y(0) | X}$ can be recovered from the marginals alone unless one imposes either the independence assumption or specifies the copula linking $Y(0)$ and $Y(1)$.
\end{proof}

Notably, the assumption of noise independence is already encoded in formal causal models like FFRCISTG~\citep{ROBINS19861393} and Pearl's nonparametric structural equation models.

\citet{melnychuk2024quantifying} recently argue that ITE distributions are fundamentally unidentifiable in the potential outcomes framework, proposing partial identification via bounds on the CDF. While this is a promising direction, we take a more pragmatic stance: we assume a specific copula, namely the independent copula, for $\varepsilon(0)$ and $\varepsilon(1)$. Although this assumption is ultimately untestable and somewhat philosophical, we believe it provides a useful starting point for further research and discussion within the community.

\subsection{Prediction intervals and violations of the noise assumption}
In the following paragraphs, we will discuss three possible characteristics of the joint distribution of the noise terms and their impact on the validity of predictive intervals and ITE distributions when assuming independence. While the true nature of these joint noise distributions is, in part, a philosophical question, it requires further attention from the research community and domain experts.

\textbf{$\mathbf{\boldsymbol{\varepsilon(0) \perp\!\!\!\perp \varepsilon(1)}}$:} In instances where the two noise distributions are mutually independent, we anticipate reasonable validity of predictive distributions with the CCT- and CMC T-learner, and, at the very least, the capability to generate conservative prediction intervals for the other CMC meta-learners.

\textbf{$\mathbf{\boldsymbol{\varepsilon(0) = \varepsilon(1)}}$:} Here, the ITE becomes deterministic, as the shared noise cancels out. This is often assumed in synthetic benchmarks. However, both CCT and CMC models will overestimate uncertainty, since they redundantly model shared aleatoric variability. A better approach would model epistemic uncertainty in both potential outcome models independently, allowing for more accurate uncertainty propagation via convolution or Monte Carlo sampling.

\textbf{{$\mathbf{\boldsymbol{\varepsilon(0)  \not\!\perp\!\!\!\perp \varepsilon(1)}}$}:} When the two noise distributions exhibit any dependency, inferring ITEs is challenging as we lack observations on the relationship between the distributions. If the noise distributions are positively correlated, potentially due to shared components in the outcome function and measurement errors in covariates, CCT- and CMC meta-learners could never guarantee valid predictive distributions. However, these models still retain the capability to yield conservatively valid prediction intervals. 

Simulation results (Figure~\ref{fig:results_setupA_B_C_D_nie_epsilon}) empirically validate these scenarios.

\newpage
\section{Proofs}
\label{ap:proofs}
\subsection{Probabilistic calibration of weighted conformal predictive system}
\label{ap:proof-wct}

\begin{defbox}
\begin{definition}[\citet{tibshiraniConformalPredictionCovariate2019a}, Definition 1]
\label{def:weighted-exchangability}
    Random variables $S_{1:n}$ are said to be weighted exchangeable, with functions $w_{1:n}$, if the density $f$ of their joint distribution can be factorized as
    \begin{equation}
        f(s_1, \ldots, s_n) = \prod_{i=1}^n w_i(s_i) g(s_1,\ldots, s_n),
    \end{equation}
    where $g$ is any function that does not depend on the ordering of its inputs, i.e., $g(s_{\sigma(1)},\ldots, s_{\sigma(n)})$ for any permutation $\sigma$ of $1, \ldots, n$.
\end{definition}
\end{defbox}

\begin{theobox}
\theoWeightedConformalTransducer*
\end{theobox}

\begin{proof}
    The proof strategy is inspired by the proof of smoothed conformal prediction (\citet{angelopoulos2024theoreticalfoundationsconformalprediction}, Lemma 9.3). \\
    Let $E_z$ denote the event that $\{Z_1,\ldots,Z_{n+1}\} = \{z_1,\ldots,z_{n+1}\}$, and let the $s_i = s(z_i;z_{1:n+1 \setminus i})$ for $i=1,\ldots,n+1$. Let $f$ be the density function of the joint sample $Z_{1:n+1}$. For each $i$,
    \begin{equation*}
        \mathbb{P}\left\{S_{n+1} = s_i | E_z\right\} = \mathbb{P}\left\{ Z_{n+1} = z_i | E_z \right\} = \frac{\sum_{\sigma | \sigma(n+1)=i} f(z_{\sigma(1):\sigma(n+1)})}{\sum_{\sigma} f(z_{\sigma(1):\sigma(n+1)})}. 
    \end{equation*}
    Since $Z_{1:n+1}$ are weighted exchangeable (Definition \ref{def:weighted-exchangability}),
    \begin{equation}
    \label{eq:weights}
        \begin{split}
            \frac{\sum_{\sigma | \sigma(n+1)=i} f(z_{\sigma(1):\sigma(n+1)})}{\sum_{\sigma} f(z_{\sigma(1):\sigma(n+1)})} & =  \frac{\sum_{\sigma|\sigma(n+1)=i} \prod_{j=1}^{n+1} w_j(z_{\sigma(j)}) g(z_{\sigma(1):\sigma(n+1)})}{\sum_{\sigma} \prod_{j=1}^{n+1} w_j(z_{\sigma(j)}) g(z_{\sigma(1):\sigma(n+1)})} \\
            & = \frac{\sum_{\sigma|\sigma(n+1)=i} \prod_{j=1}^{n+1} w_j(z_{\sigma(j)}) g(z_{1:n+1})}{\sum_{\sigma} \prod_{j=1}^{n+1} w_j(z_{\sigma(j)}) g(z_{1:n+1})} \\
            & = \frac{\sum_{\sigma|\sigma(n+1)=i} \prod_{j=1}^{n+1} w_j(z_{\sigma(j)})}{\sum_{\sigma} \prod_{j=1}^{n+1} w_j(z_{\sigma(j)})} \\ 
            & \stackrel{(*)}{=} \frac{w(x_i)}{\sum_{j=1}^{n+1} w(x_j)} = p_i^w
        \end{split}
    \end{equation}
    The simplification at $(*)$ can be made because we consider only a special case of weighted exchangeability, the covariate shift, where $w_{1:n} \equiv 1$ and $w_{n+1}((x,y))=w(x)=\frac{d\tilde{P}_X(x)}{dP(x)}$. Nevertheless, our findings can be readily applied to the general concept of weighted exchangeability.
    \\\\
    The result in \ref{eq:weights} shows that,
    \begin{equation}
    \label{eq:conditional-dist-on-observed-seq}
        S_{n+1} | E_z \sim \sum_{i=1}^{n+1} p_i^w \delta_{s_i}.
    \end{equation}
    For each $j \in [1:n+1]$ define,
    \begin{equation}
        p_j(\phi) = \sum_{i=1}^{n+1} [S_i < S_{j}^y] p_i^w(X_{j}) + \sum_{i=1}^{n+1} [S_i^y = S_{j}^y] p_i^w(X_{j}) \phi.
    \end{equation}
    where $S_i = S_i^{Y_{n+1}}$ for any $i \in [1:n+1]$. \\\\
    By construction, the result in \ref{eq:conditional-dist-on-observed-seq}, and the tower law,
    \begin{equation}
    \label{eq:tower-expectation}
        \mathbb{P}(p_{n+1}(\phi) \leq \alpha) = \sum_{j=1}^{n+1} p_j^w \mathbb{P}(p_j(\phi) \leq \alpha) = \mathbb{E}\left[ \sum_{j=1}^{n+1} p_j^w \mathbb{P}(p_j(\phi) \leq \alpha | S_{1:n+1}) \right].
    \end{equation}
    Define the weighted quantile,
    \begin{equation*}
        q = \text{Quantile}\left(\sum_{i=1}^{n+1} p_i^w(X_i) \delta_{S_i}; 1-\alpha\right),
    \end{equation*}
    and let
    \begin{equation*}
        N^{-} = \sum_{j=1}^{n+1} [S_j < q] p_j^w,\qquad   N^{=} = \sum_{j=1}^{n+1} [S_j = q] p_j^w.
    \end{equation*}
    By definition of the weighted quantile, for any $j$ if $S_j = q$,
    \begin{equation*}
        p_j({\phi}) = N^{-} + \phi N^{=},
    \end{equation*}
    and thus,
    \begin{equation*}
         p_j({\phi}) \leq \alpha \iff \phi \leq \frac{\alpha - N^{-}}{N^{=}}
    \end{equation*}
    Therefore,
    \begin{equation}
        \mathbb{P}(p_j(\phi) \leq \alpha | S_{1:n+1}) = [S_j < q] + [S_j = q] \frac{\alpha - N^{-}}{N^{=}}.
    \end{equation}
    Hence, combined with the results of \ref{eq:tower-expectation},
    \begin{equation}
        \begin{split}
            \mathbb{P}(p_j(\phi) \leq \alpha) &= \mathbb{E}\left[ \sum_{j=1}^{n+1} p_j^w \mathbb{P}(p_j(\phi) \leq \alpha | S_{1:n+1}) \right] \\
            &= \mathbb{E}\left[ \sum_{j=1}^{n+1} p_j^w \left\{ [S_j < q] + [S_j = q] \frac{\alpha - N^{-}}{N^{=}} \right\} \right] \\
            &= \mathbb{E}\left[ \sum_{j=1}^{n+1} [S_j < q] p_j^w + \sum_{j=1}^{n+1} [S_j = q] p_j^w \frac{\alpha - N^{-}}{N^{=}} \right] \\
            &= \mathbb{E}\left[ N^{-} + N^{=} \frac{\alpha - N^{-}}{N^{=}} \right] = \mathbb{E}\left[ \alpha \right] = \alpha.
        \end{split}
    \end{equation}
    Thus, the distribution of $Q\left(X_{n+1}, Y_{n+1}, \phi, \frac{d\tilde{P}}{dP};Z_{1:n}\right)$ is uniform, completing the proof.
\end{proof}

\subsection{Probabilistic calibration of CCT-learner}
\label{ap:proof-cct-validity}

\begin{theobox}
\cctCalibration*
\end{theobox}
\begin{proof}
    The proof follows directly from Theorem \ref{theo:valid-wct}, where we assume weighted exchangeability according to,
    \begin{equation}
    Y_{n+1}(1) - Y_{n+1}(0)|E_{q_{Y(1)} - q_{Y(0)}} \sim \sum_{i=1}^{n_0} \sum_{j=1}^{n_1}  p^{w_0}_i p^{w_j}_i \delta_{q_{Y(1),j} - q_{Y(0),i}} + [1 - \{1-p_{n+1}^{w_1}\}\{1-p_{n+1}^{w_0}\}] \delta_{\infty}.
\end{equation}
\end{proof}

\subsection{Existence universal consistent CCT- and CMC T-learner}
\label{ap:proof-universal}
The following section presents and proves Theorem \ref{theo:consistency}, which states the existence of a universal consistent CCT- and CMC T-learner, i.e., the conformal predictive distribution approaches the true data generating distribution as $n \rightarrow \infty$ under any data-generating distribution. We introduce here Definition \ref{def:consistent}, taken in verbatim from \cite{vovkUniversalPredictiveSystems2022}, which lays out our notion of consistency for CPS.

\begin{definition}[in verbatim \citet{vovkUniversalPredictiveSystems2022}]
\label{def:consistent}
A conformal predictive system $Q$ is consistent for a probability measure $\mathcal{P}$ on $Z := X \times \mathbb{R}$ if, for any bounded continuous function $f: \mathbb{R} \rightarrow \mathbb{R}$,
\begin{equation}
    \int f dQ_n - \mathbb{E}_{\mathcal{P}}(f | X_{n+1}) \rightarrow 0 \qquad (n \rightarrow \infty)
\end{equation}
in probability, where:
\begin{itemize}
    \item $Q_n$ is the predictive distribution $Q_n: y \mapsto Q(X_{n+1}, \phi; Z1:n)$ output by Q as its forecast for the label $Y_{n+1}$ corresponding to the test object $X_{n+1}$ based on the training data $Z_{1:n}$, where $Z_i = (X_i,Y_i)$ for all $i$.
    \item $\mathbb{E}_{\mathcal{P}}(f | X_{n+1})$ is the conditional expectation of $f(y)$ given $x = X_{n+1}$ under $(x,y) \sim \mathcal{P}$.
    \item $Z_i = (X_i, Y_i) \sim \mathcal{P}, i=1,...,n$, $Z_{n+1} \sim \mathcal{P}$, and $\phi \sim Uniform(0,1)$, are assumed all independent.
\end{itemize}
\end{definition}

\begin{theobox}
    \theoConsistency*
\end{theobox}

\begin{proof}
The philosophy behind the MC simulations in the CMC T-learner is to not perform a convolution of both outcome predictive distributions, as this would quickly become an expensive operation that would need to be performed for every example during inference. However, to prove that there exists a universal consistent learner, we replace this MC sampling in Algorithm \ref{alg:fit_cmc} with the convolution operation,
\begin{equation}
    \begin{split}
        Q_{ITE}&(X_{n+1},y,\phi, \pi; Z_{1:n}) = \\
        &\int Q_{Y(1)}(X_{n+1},y+t,\phi, w_1=\frac{1}{\pi};Z_{1:n1}^{W=1})dQ_{Y(0)}(X_{n+1}, t,\phi,\frac{1}{1-\pi};Z_{1:n0}^{W=0}).
    \end{split}
\end{equation}
Thus, becoming a CCT-learner is equivalent to the CMC T-learner approach asymptotically.

Considering the existence of a universal consistent learner, we no longer need to worry about the covariate shift. We can replace the WCPS with CPS in the CCT- and CMC T-learner. This is equivalent to setting the propensity score to $0.5$ for every covariate.

By Theorem 3 in \cite{vovkUniversalPredictiveSystems2022}, we know that there exists universal CPS if the object space $\mathcal{X}$ is standard Borel, meaning that for any probability measure $\mathcal{P}$ on $Z$, the CPS is consistent. Consequently, a universal CPS exists that can model the true data-generating distribution outcome under treatment $Y(1)$ and control $Y(0)$. These distributions are weakly converging distributions. 
Since we assume that the predictive distributions of the outcome are independent,
\begin{equation}
    Y(0) \perp\!\!\!\perp Y(1) | X,
\end{equation}
the outcome variables are independent conditionally on the covariate set $X$. Then, by the definition of convolution and the fact that $ITE = Y(1) - Y(0)$, the convolution of the true conditional cumulative distribution of outcome under treatment $F_{Y(1)|X}$ and outcome under negative control $F_{-Y(0)|X}$ equals the true conditional cumulative ITE distribution,
\begin{equation}
    F_{ITE|X}(y) = \int F_{Y(1)|X}(y+t) dF_{Y(0)|X}(t).
\end{equation}
Since the convolution of two independent weakly converging sequences converges weakly to the independent joint distribution (Theorem 105, \citet{mcleishSTAT901Probability2005}), a universal consistent CCT-learner exists and consequently also a CMC T-learner.
\end{proof}

\clearpage
\section{Experiments results}
\label{ap:exp_results}
\begin{figure}[H]
    \centering
    \includegraphics[width=\linewidth]{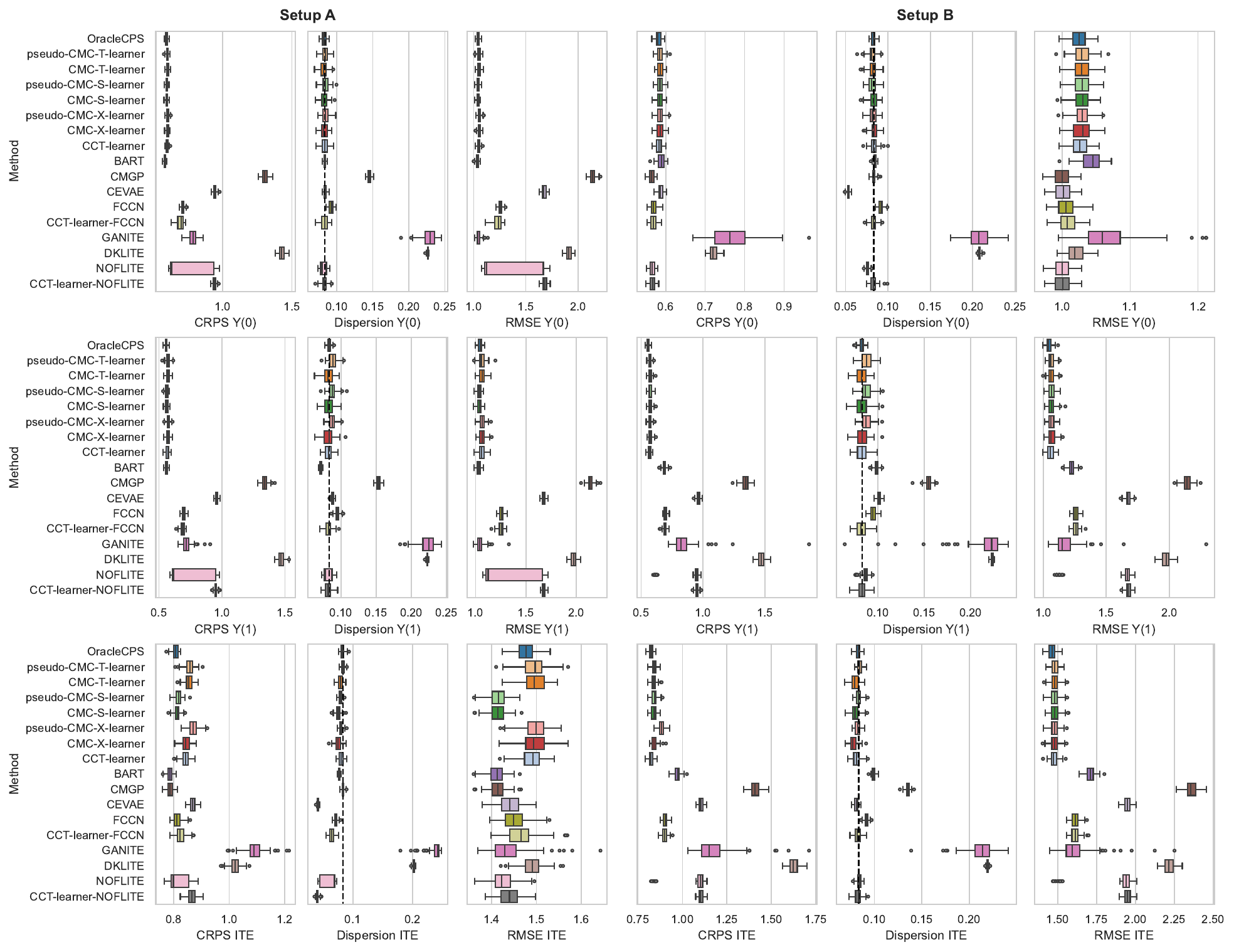}
    \caption{Simulation results for synthetic experiment setups A and B from \citet{alaaConformalMetalearnersPredictive2023}.}
    \label{fig:alaa-dist}
    \vspace*{\fill}
\end{figure}

\clearpage
\begin{figure}[H]
    \centering
    \includegraphics[width=\linewidth]{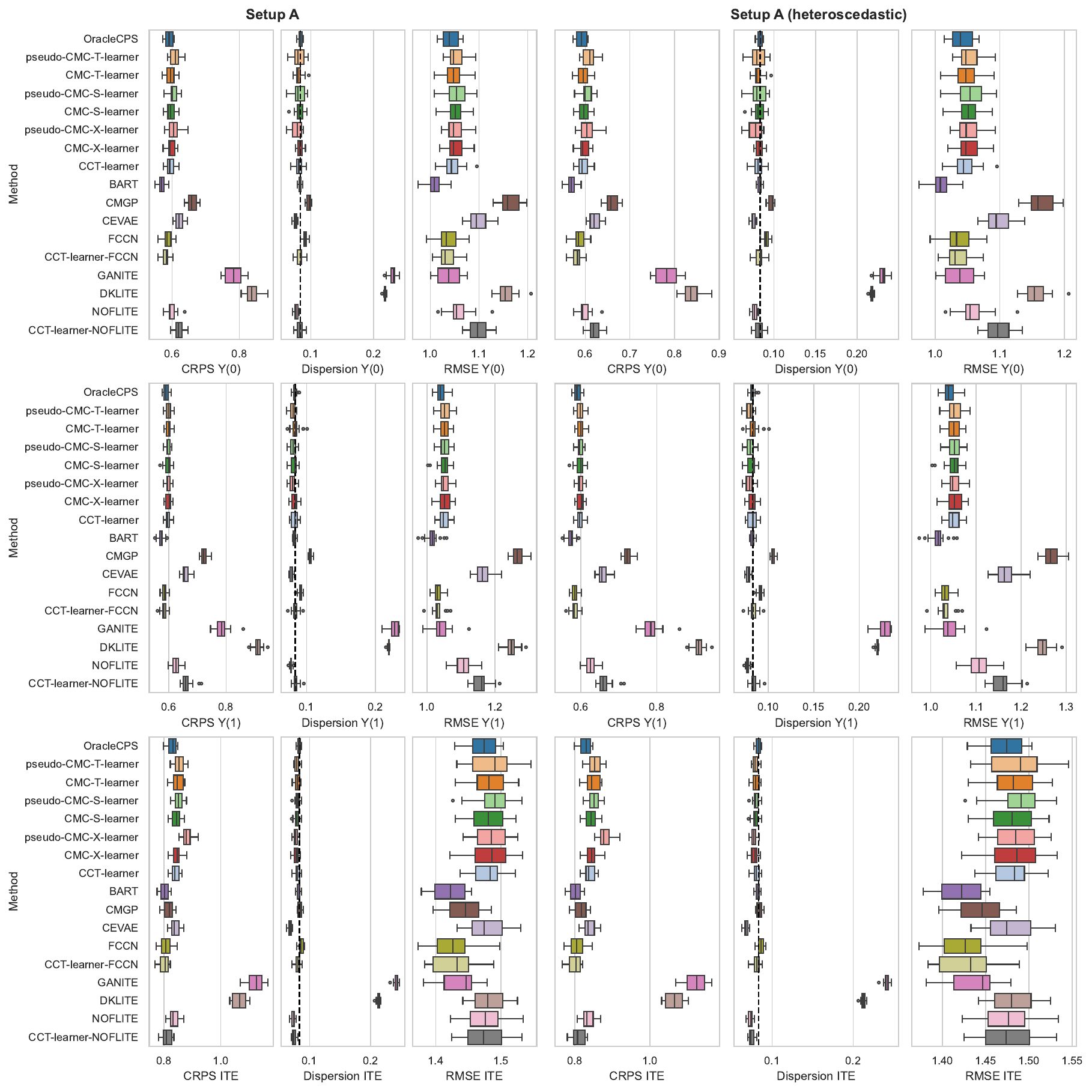}
    \caption{Simulation results for synthetic experiment setup A \citet{nieQuasioracleEstimationHeterogeneous2021}.}
    \label{fig:nw-A-dist}
\end{figure}

\begin{figure}[H]
    \centering
    \includegraphics[width=\linewidth]{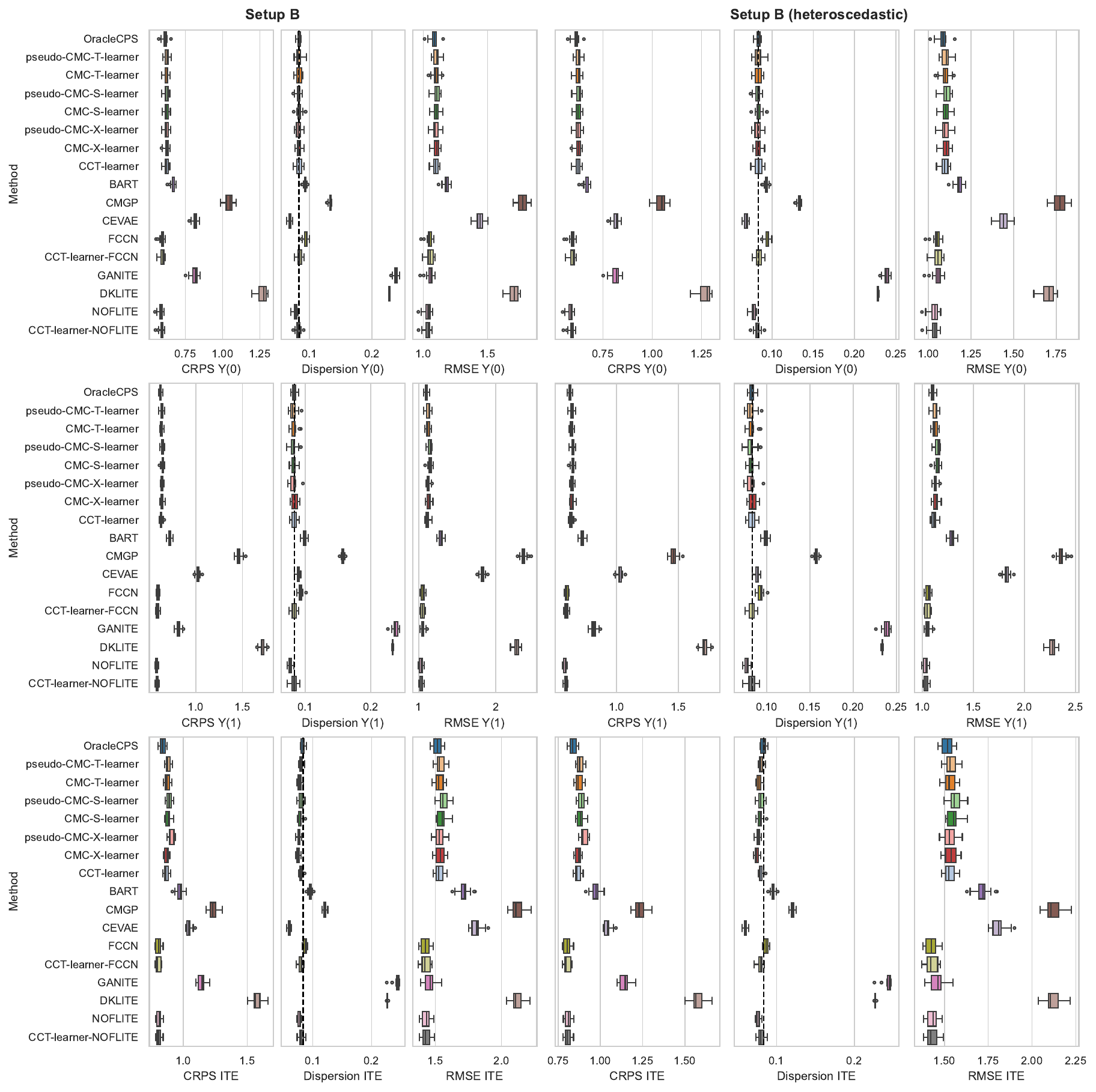}
    \caption{Simulation results for synthetic experiment setup B \citet{nieQuasioracleEstimationHeterogeneous2021}.}
    \label{fig:nw-B-dist}
\end{figure}

\begin{figure}[H]
    \centering
    \includegraphics[width=\linewidth]{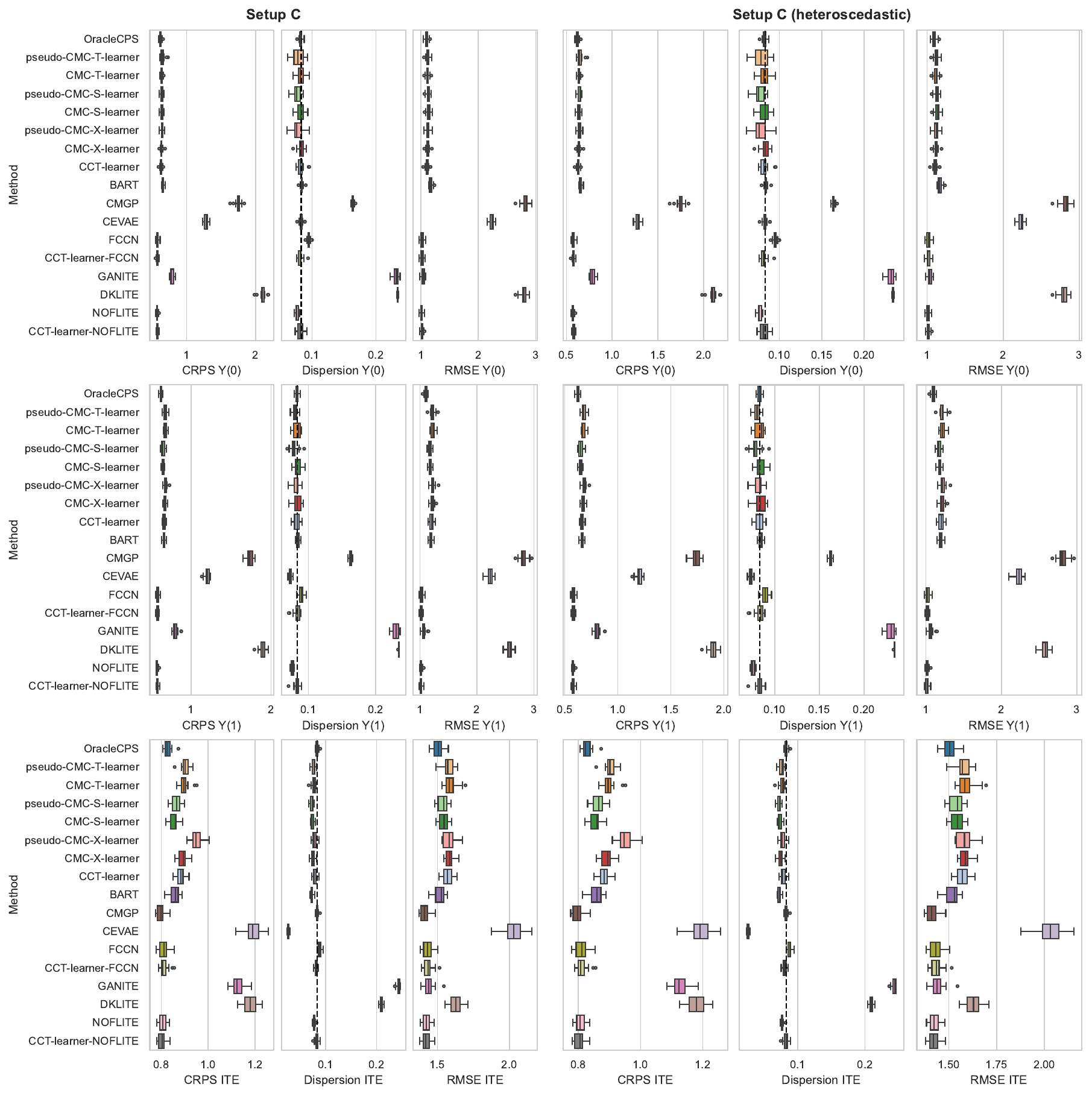}
    \caption{Simulation results for synthetic experiment setup C \citet{nieQuasioracleEstimationHeterogeneous2021}.}
    \label{fig:nw-C-dist}
\end{figure}

\begin{figure}[H]
    \centering
    \includegraphics[width=\linewidth]{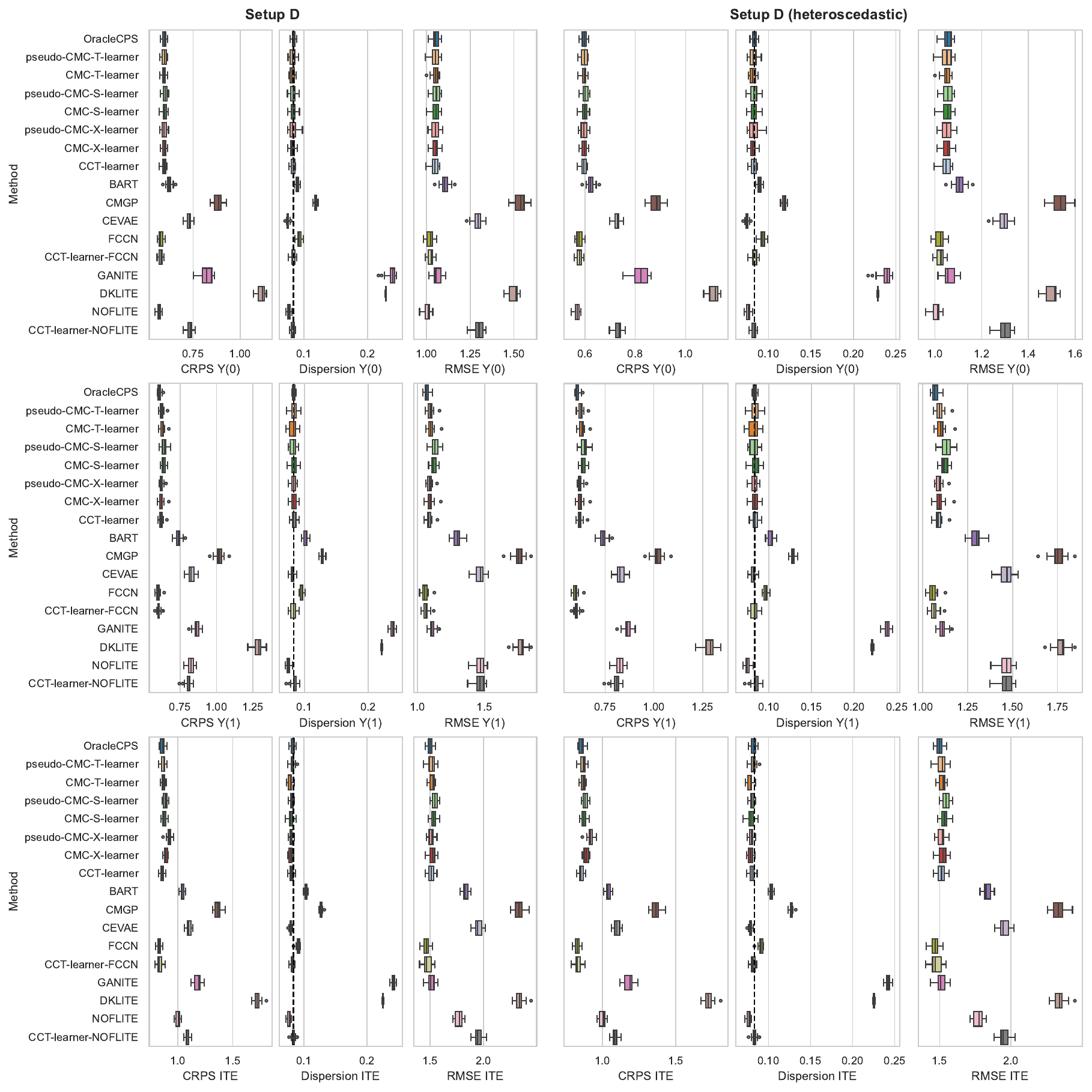}
    \caption{Simulation results for synthetic experiment setup D \citet{nieQuasioracleEstimationHeterogeneous2021}.}
    \label{fig:nw-D-dist}
\end{figure}

\begin{figure}[H]
    \centering
    \includegraphics[width=\linewidth]{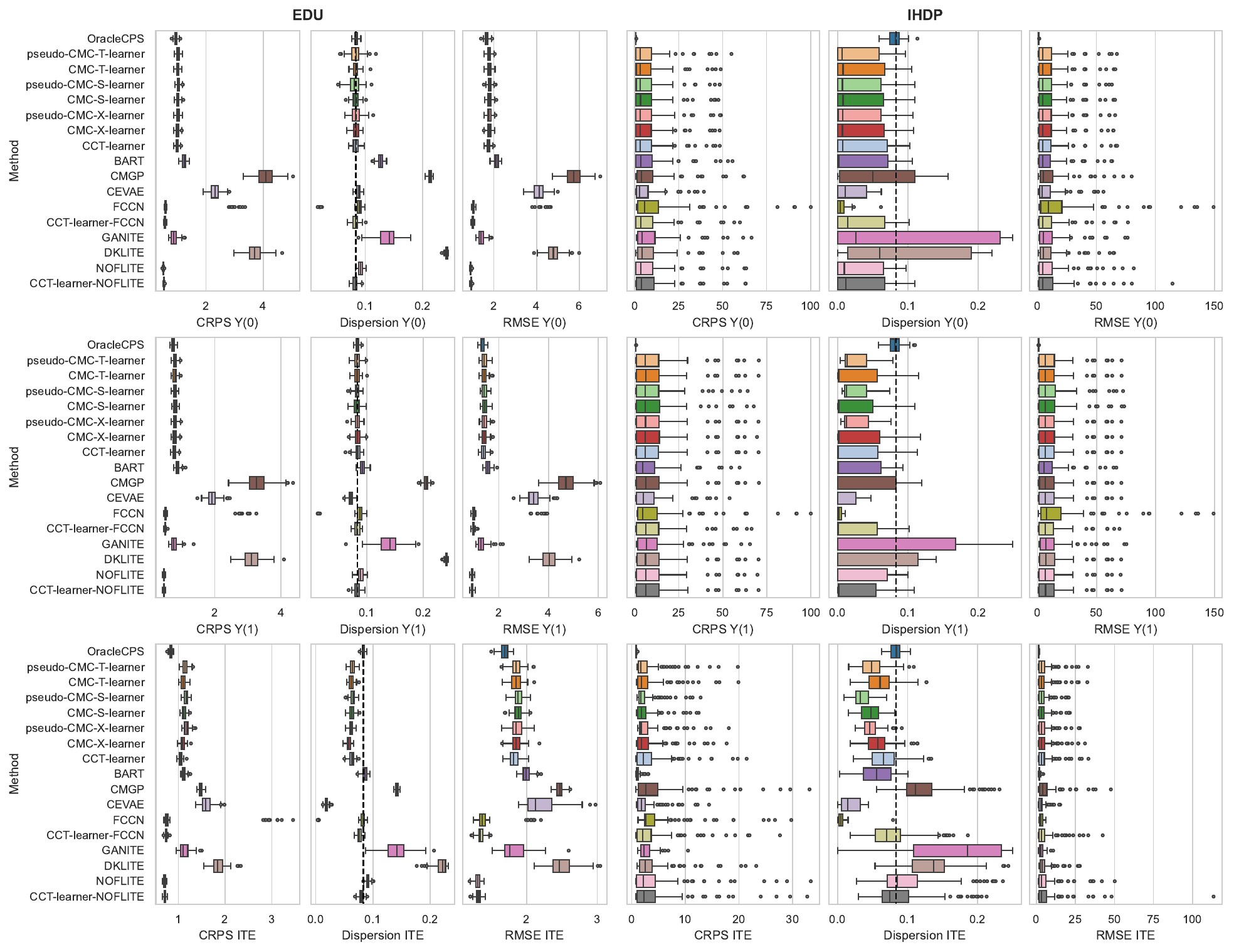}
    \caption{Simulation results for semi-synthetic experiments EDU and IHDP.}
    \label{fig:edu-ihdp-dist}
\end{figure}

\begin{figure}[H]
    \centering
    \includegraphics[width=\linewidth]{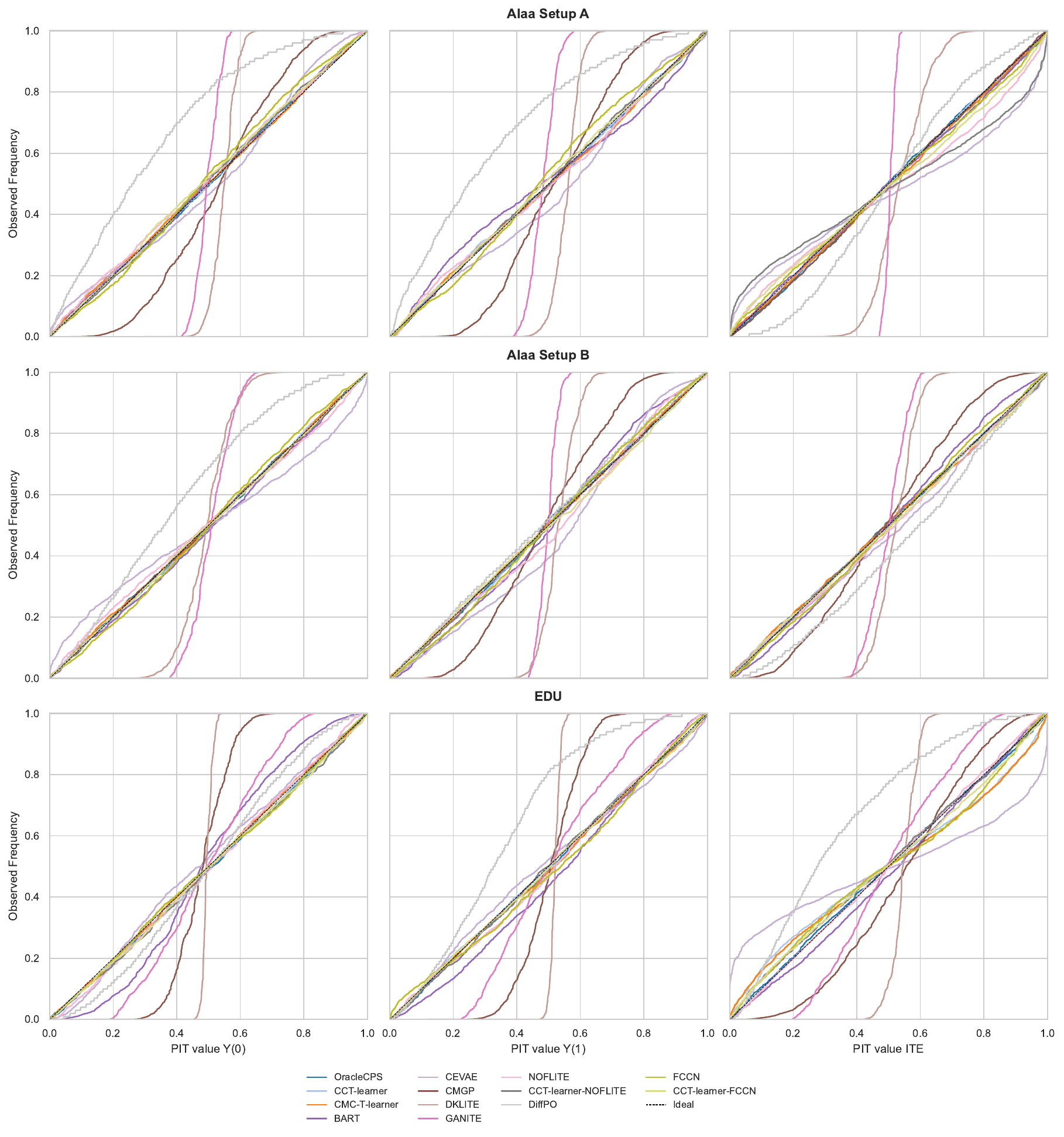}
    \caption{Observed frequency of PIT-values on the synthetic experiments of \citet{alaaConformalMetalearnersPredictive2023} and the semi-synthetic experiment EDU.} 
    \label{fig:pit-values}
\end{figure}

\subsection{Conformal prediction intervals}
Both CCT- and CMC meta-learners return conditional predictive distributions of the ITE. Prediction intervals of confidence level $1-\alpha$ are formed by taking the $\frac{\alpha}{2}$ and $1-\frac{\alpha}{2}$ quantiles from the predictive distribution as the lower and upper bounds, respectively. Our method produces symmetric, two-sided intervals with equally distributed probability density. In contrast, the approaches by \citet{leiConformalInferenceCounterfactuals2021} and \citet{alaaConformalMetalearnersPredictive2023} generate asymmetric intervals, offering an efficiency advantage due to smaller interval lengths~\citep{romanoConformalizedQuantileRegression2019}. However, these asymmetric intervals can be misleading in practical applications as they might intuitively imply equal probability density on both sides of the interval.

The experiments assessed coverage and efficiency of prediction intervals of the CCT-learner and (pseudo) CMC T-learner against WCP (exact, inexact, and naive)~\citep{leiConformalInferenceCounterfactuals2021}, CM-Learner (DR)~\citep{alaaConformalMetalearnersPredictive2023}, and a CPS Oracle fitted as a standard regression problem on the ITE.

\begin{figure}
  \centering
  \includegraphics[width=\columnwidth]{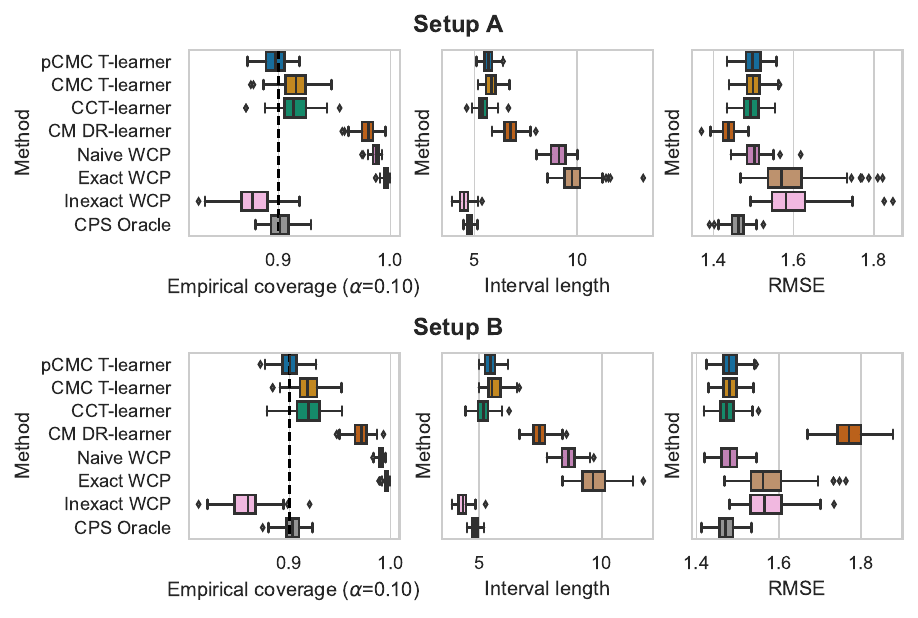}
  \caption{Simulation results for setups A and B from \citet{alaaConformalMetalearnersPredictive2023} for 90\% ITE prediction interval.}
  \label{fig:results_setupA_B_90_alaa}
\end{figure}
\begin{figure}
  \centering
  \includegraphics[width=\textwidth]{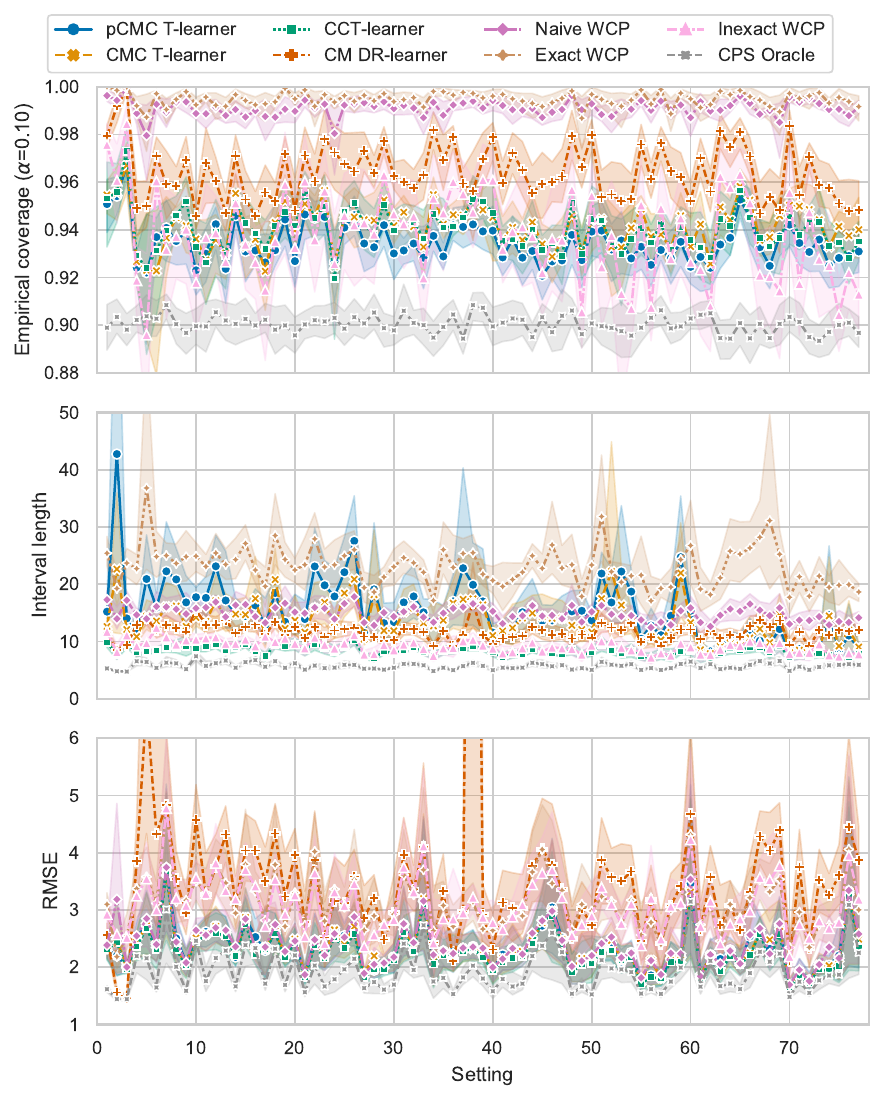}
  \caption{Results on 77 settings of the ACIC2016 dataset for the 90\% ITE prediction interval.}
  \label{fig:results_acic2016}
\end{figure}
\begin{figure*}[!ht]
    \centering
    \includegraphics[width= \linewidth]{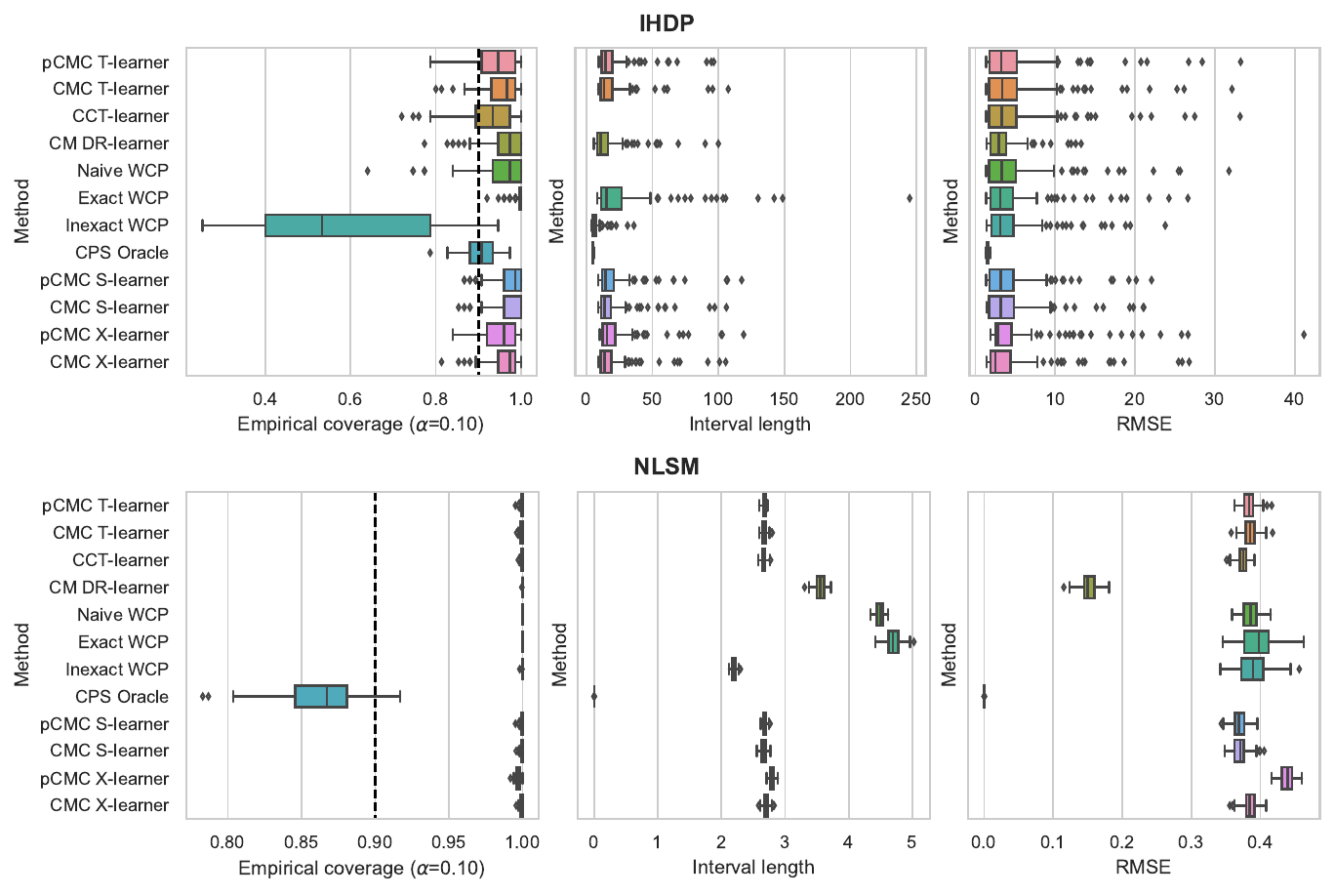}
    \caption{Simulation results for estimating 90\% ITE prediction interval on the IHDP and NLSM semi-synthetic datasets.}
    \label{fig:results_90_ihdp_nlsm}
\end{figure*}

\paragraph{Coverage and efficiency of prediction intervals}
On synthetic datasets, the CCT- and CMC meta-learners consistently demonstrated slightly conservative coverage across various simulations while maintaining high efficiency, indicated by small average interval widths. In contrast, the inexact WCP, while more efficient in some cases, often did so at the cost of significant under coverage. The CM DR learner was consistently more conservative than our proposed methods. Other WCP approaches, with proven conservative validity, were extremely conservative across all confidence levels and experiments, likely due to their over-conservative design, employing either Bonferroni correction or double conformal inference procedure~\citep{leiConformalInferenceCounterfactuals2021}.
\\
Comparing the CCT- and (p)CMC T-learners, both showed similar coverage levels across all experiments. The pCMC T-learner produced less conservative intervals than the other two approaches; however, this did not necessarily translate to greater efficiency compared to the CMC T-learner. The CCT-learner consistently produced more efficient intervals across all datasets, possibly because it can use more samples for the potential outcome models as the (p)CMC learners require an additional data split. Applying a cross-fitting procedure might mitigate this effect for the (p)CMC learners, a possibility we leave for future work. Additionally, the pCMC showed slightly greater computational efficiency, needing one less variable per MC sample than the CMC learner.
\\
Within the CMC meta-learners, performance differences were small and dependent on the data-generating process. For example, in setup D from \citet{nieQuasioracleEstimationHeterogeneous2021}, where treatment and control groups have unrelated outcomes, the T-learner was slightly more efficient than the CMC S- and X-learners, possibly due to the absence of synergies for jointly learning both outcomes.
\\
For the three semi-synthetic datasets, we performed experiments with covariates from real-world data. As here no propensity scores are provided, these resemble a realistic scenario. We estimate the propensity scores using logistic regression. The most extensive experiment was conducted on the 77 settings of the ACIC2016 dataset~\cite{dorie_automated_2019}, which contain 58 covariates (3 categorical, 5 binary, 27 counts, 23 continuous) with varying complexity of outcome surfaces, degree of confounding, overlap, and treatment effect heterogeneity. Results, depicted in Figure~\ref{fig:results_acic2016}, show that our proposed approaches provide conservatively valid prediction intervals across all settings, with significant conservative coverage in most settings, though less than the CM DR-learner and valid WCP approaches, likely due to violations in the assumption of conditional independence of the noise distributions. Regarding efficiency, the CCT-learner performed best for the reasons previously mentioned. As can be seen in Figure~\ref{fig:results_90_ihdp_nlsm}, a similar conclusion can be drawn for the other semi-synthetic experiments.

\begin{figure}
  \centering
  \includegraphics[width=0.7\textwidth]{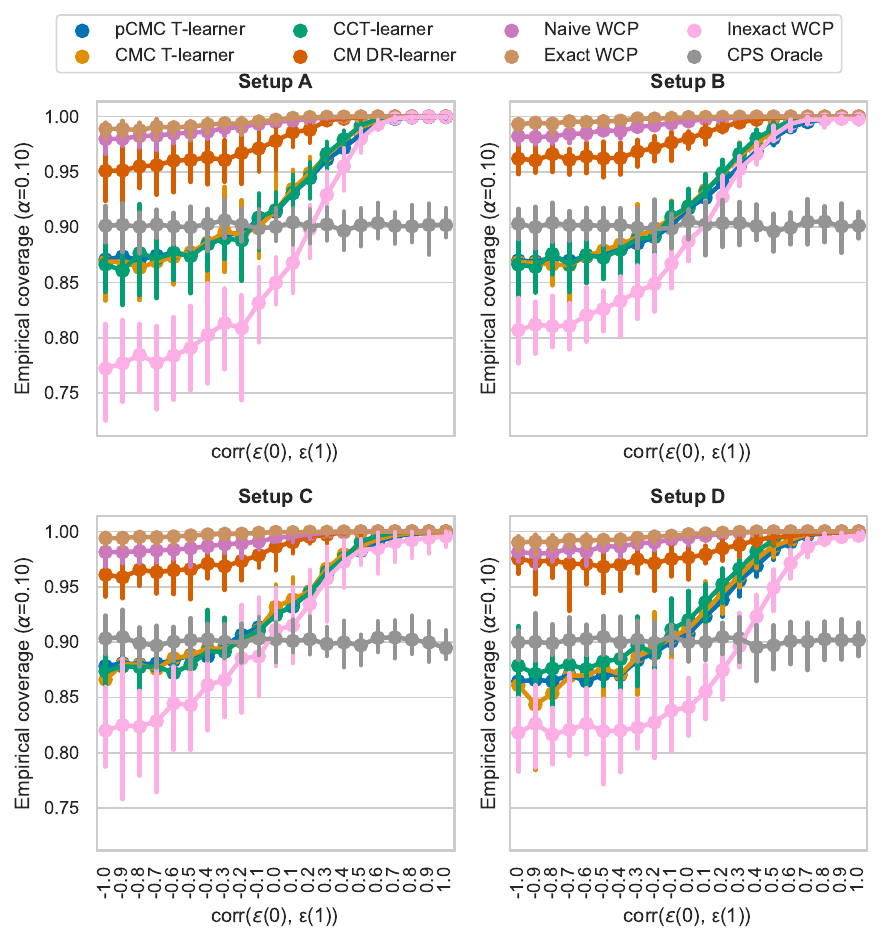}
  \caption{Simulation results for setups A, B, C, and D from \citet{nieQuasioracleEstimationHeterogeneous2021}, with increasing correlation between the noise terms of potential outcome under treatment $Y^1$ and control $Y^0$. The vertical lines represent the variation in the different simulation runs and cover an interval with the results of 95\% of the runs.}
  \label{fig:results_setupA_B_C_D_nie_epsilon}
\end{figure}

\paragraph{Performance under different relations between noise distributions}
The relationship between the noise distributions of the two potential outcomes significantly affects coverage and the validity of interval predictions and predictive distributions, as evidenced in our experiments where we incrementally increased correlations between the two noise terms. Figure~\ref{fig:results_setupA_B_C_D_nie_epsilon} shows that all methods exhibit more overconfidence as correlation increases starting from zero. When correlation becomes negative, both the CCT- and CMC-learners lose some validity, though not as much as the inexact WCP. The CM DR-learner and valid WCP approaches, while experiencing decreased coverage, remain valid. While this is one of the limitations of CCT- and CMC-learners, as this is an untestable condition, one could argue if such a scenario happens in reality.

\begin{figure}[!ht]
      \centering
      \includegraphics[width=\textwidth]{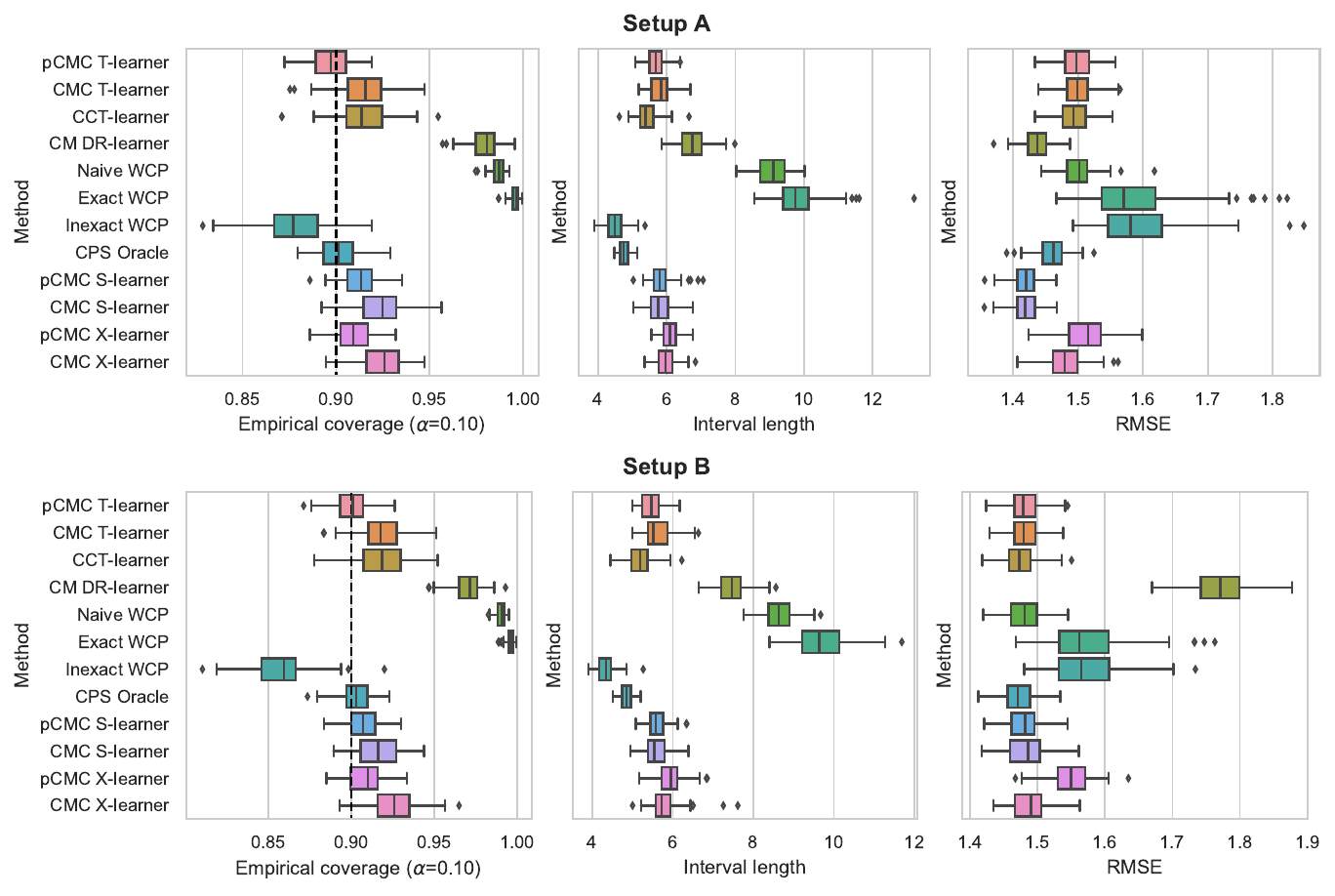}
      \caption{Simulation results for setups A and B from \citet{alaaConformalMetalearnersPredictive2023} for 90\% ITE prediction interval. (With other CMC meta-learners)}
\end{figure}
\begin{figure}
    \centering
    \includegraphics[width= 0.55\linewidth]{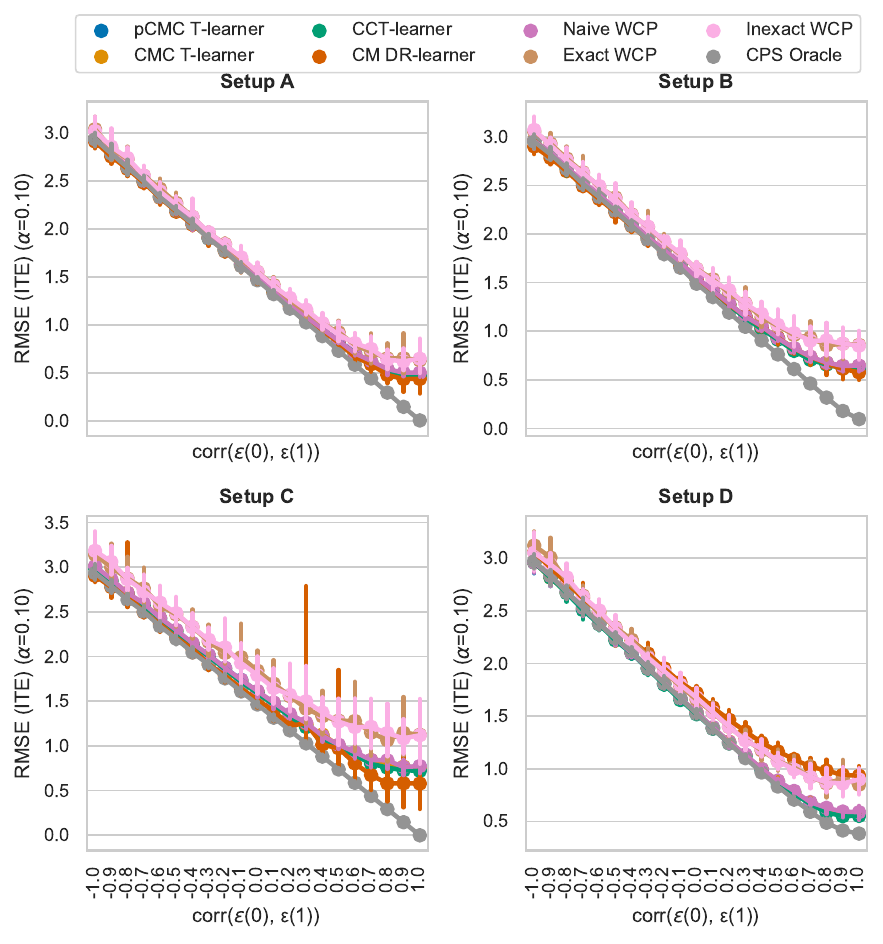}
    \caption{Simulation results for setups A, B, C, and D from \citet{nieQuasioracleEstimationHeterogeneous2021}, with increasing correlation between the noise terms of potential outcome under treatment $Y^1$ and control $Y^0$. The vertical lines represent the variation in the different simulation runs and cover an interval with the results of 95\% of the runs.}
    \label{fig:results_setupA_B_C_D_nie_epsilon_rmse}
\end{figure}

\begin{figure}
    \centering
    \includegraphics[width= 0.55\linewidth]{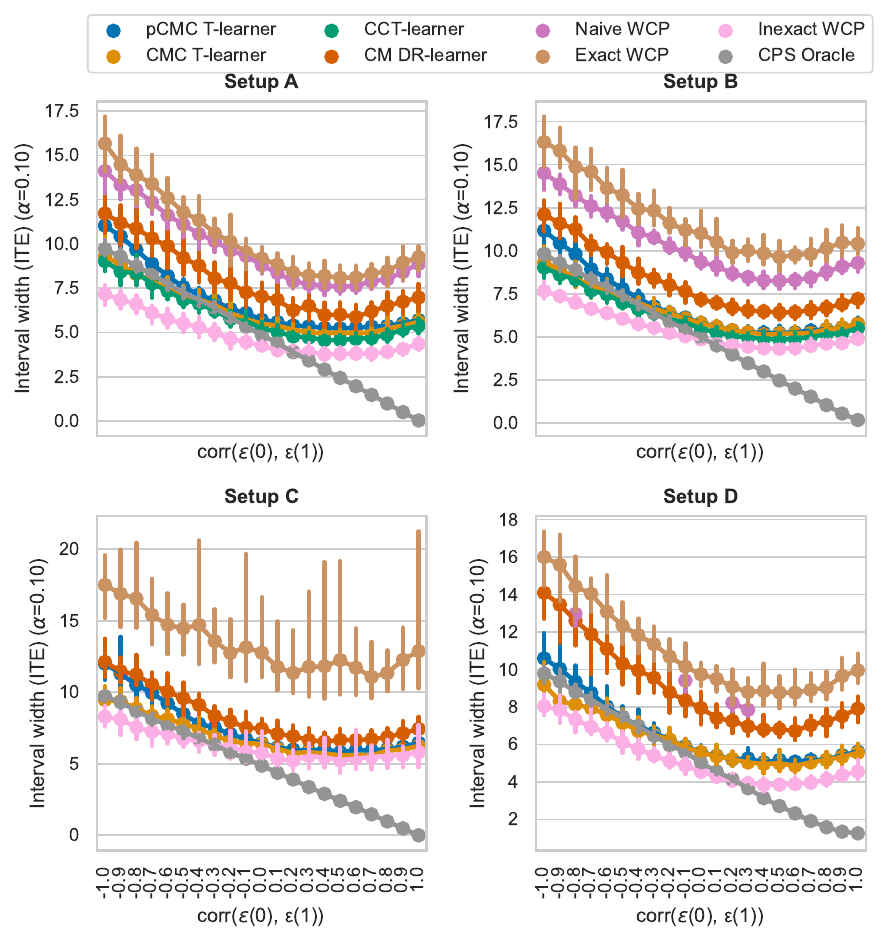}
    \caption{Simulation results for setups A, B, C, and D from \citet{nieQuasioracleEstimationHeterogeneous2021}, with changing correlation between the noise terms of potential outcome under treatment $Y^1$ and control $Y^0$. The vertical lines represent the variation in the different simulation runs and cover an interval with the results of 95\% of the runs.}
    \label{fig:results_setupA_B_C_D_nie_epsilon_efficiency}
\end{figure}

\begin{figure*}[!ht]
    \centering
    \includegraphics[width=0.98\textwidth]{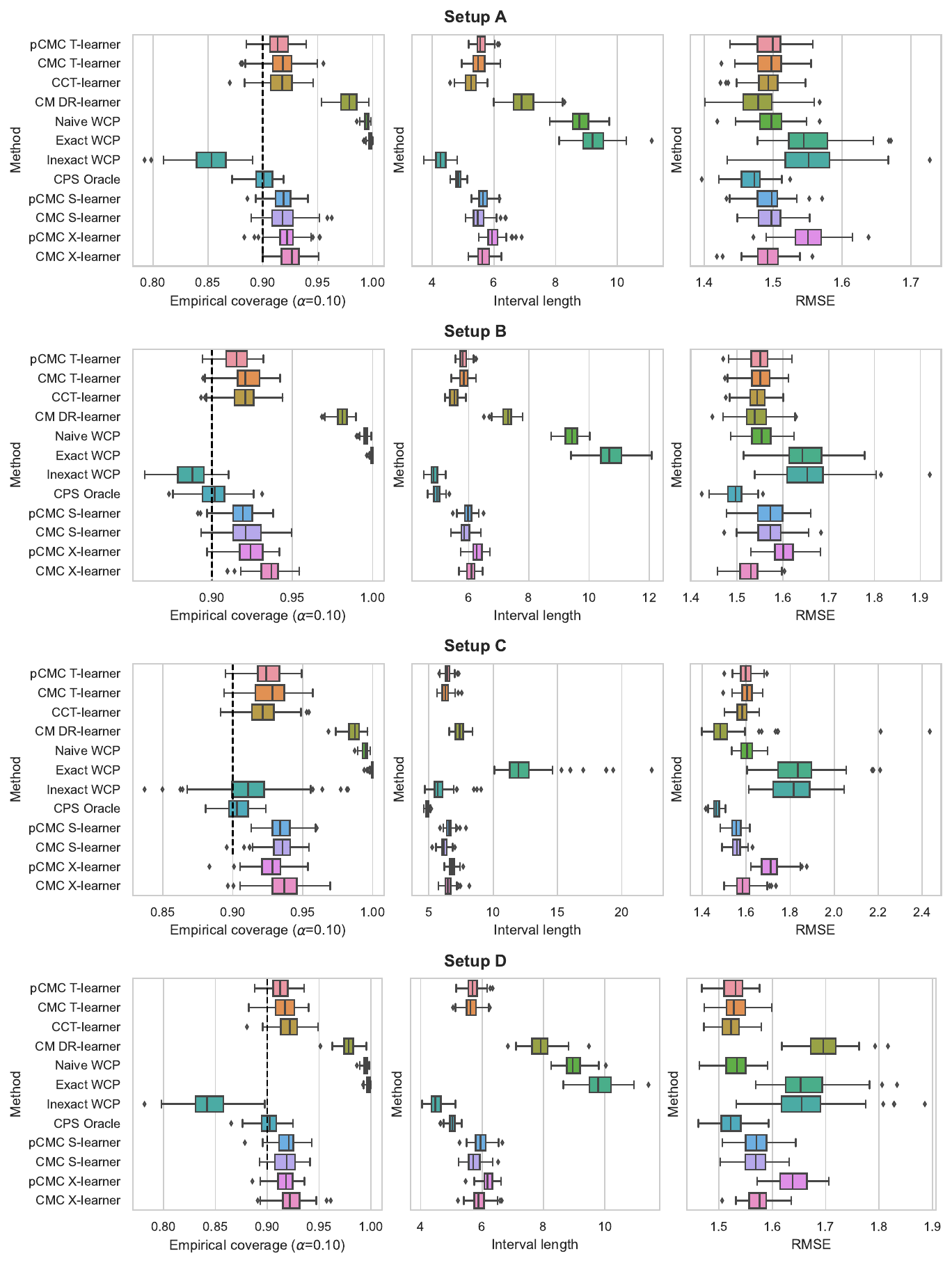}
    \caption{Simulation results for setups A, B, C, and D from \citet{nieQuasioracleEstimationHeterogeneous2021} for estimating 90\% ITE prediction interval.}
    \label{fig:results_setupA_B_C_D_90_nie}
\end{figure*}

\clearpage
\section{Experimental setups}
\label{ap:exp-setup}

\subsection{Experiments compute resources}
\label{ap:exp-resources}
All experiments were conducted on a private compute cluster. Each experimental run utilized 4 CPU cores and 30 GB of memory, with no GPU acceleration required. For experiments involving random forest-based methods, individual runs completed in under 30 seconds, while other configurations (e.g., preprocessing, hyperparameter tuning) exhibited longer execution times proportional to dataset size and algorithmic complexity.

\subsection{Synthetic experiments}
\label{sec:synth-exp}
\subsubsection{Alaa et al. synthetic data experiments}\label{body:alaa_synthetic}
The first synthetic data experiments follows the setup proposed by \citet{alaaConformalMetalearnersPredictive2023}. Two setups are used: A and B. Each setup uses 5000 samples and 10 covariates. The experiments are performed using 100 simulations, each using a different random seed and using the $\alpha$ value of $0.10$. Additionally, for our proposed model, we evaluate the probabilistic calibration of the predictive distribution. The CCT and CMC models use normalized split WCPS. The results are shown in Figure \ref{fig:alaa-dist}, \ref{fig:pit-values}, and \ref{fig:results_setupA_B_90_alaa}.

\paragraph{Data generating process}
This data generation process entails both Setup A and Setup B. For Setup A, where $\gamma=1$, represents an experiment where the treatment has no effect, while for Setup B, where $\gamma=0$, represents a heterogeneous treatment effect. X is sampled from a uniform distribution, $d$ is the number of covariates, $\tau$ represents the treatment effect, $\gamma$ is the treatment heterogeneity, $\pi$ is the propensity, $\varepsilon(0)$ and $\varepsilon(1)$ are Gaussian noise terms, and $W$ is the treatment assignment.

\begin{equation}
    X \sim Uniform([0,1],size = (n,d)) 
\end{equation}
\begin{equation}
    \tau = \frac{2}{1+exp(-12(X_0-0.5))}\frac{2}{1+exp(-12(X_1-0.5))}
\end{equation}
\begin{equation}
    \tau_0 = \gamma  \tau
\end{equation}
The propensity $\pi$ is sampled from a Beta distribution with shape parameters 2 and 4.
\begin{equation}
    \pi = \frac{1+Beta(X_0,2,4)}{4}
\end{equation}
\begin{equation}
    \varepsilon(1) \sim \mathcal{N}(0,1)
\end{equation}
\begin{equation}
    \varepsilon(0) \sim \mathcal{N}(0,1)
\end{equation}
\begin{equation}
    Y(0) = \tau_0 + \varepsilon(0)
\end{equation}
\begin{equation}
    Y(1) = \tau + \varepsilon(1)
\end{equation}
The treatment is sampled from a uniform distribution between 0 and 1. If the value is below the propensity the assignment is \textit{treatment}, otherwise \textit{control}.
\begin{equation}
    W \sim (Uniform(0,1) < \pi)
\end{equation}
\begin{equation}
    y = W  Y(1) + (1-W)  Y(0)
\end{equation}

\subsubsection{Nie and Wager synthetic data experiments}\label{body:nie_and_wager_exp}
The second synthetic data experiments use the setups of \citet{nieQuasioracleEstimationHeterogeneous2021}. Four setups are used: A, B, C, and D. Each setup uses 5000 samples and is split into a training and testing set using a 50\% split. Every setup is simulated with uncorrelated noise terms, uses five covariates, and a standard deviation of the error term of 1. Additionally, we make for each setup a heteroscedastic setting that makes the variance of the noise distribution dependent on the covariates, $\sigma = \sigma_{het}$. The same approach is taken here as in Section~\ref{body:alaa_synthetic} with 100 simulations, each using different random seeds and using the same $\alpha$ values. The CCT and CMC models also use normalized split WCPS. The results are shown in Figure \ref{fig:nw-A-dist}, \ref{fig:nw-B-dist}, \ref{fig:nw-C-dist}, \ref{fig:nw-D-dist}, and \ref{fig:results_setupA_B_C_D_90_nie}.

Besides the aforementioned experiments with the data-generating processes proposed by \citet{nieQuasioracleEstimationHeterogeneous2021}, we also perform a noise dependency impact experiment with it. This experiment will investigate the impact of changing the dependency between the noise variables, $\varepsilon(0)$ and $\varepsilon(1)$, of $Y(0)$ and $Y(1)$, respectively, on the different conformal frameworks. The only difference with the previous experiment is that the noise correlation coefficient $c$ changes from $-1$ to $1$ in increments of $0.1$. The results are shown in Figure~\ref{fig:results_setupA_B_C_D_nie_epsilon}, \ref{fig:results_setupA_B_C_D_nie_epsilon_efficiency}, and \ref{fig:results_setupA_B_C_D_nie_epsilon_rmse}.

\paragraph{Data generating process: Setup A} 
Setup A simulates a difficult nuisance component and an easy treatment effect. $d$ is the number of covariates and $n$ the number of samples. $X$ is created as an independent variable, $n$ rows and $d$ columns thus consisting of $[X_0, X_1, ..., X_{d-1}]$ each with length $n$. The other variables are the expected outcome $b$, the propensity of receiving treatment $\pi$, $\tau$ the CATE, $W$ the treatment assignment, $Y(0)$ outcome for control, $Y(1)$ for treatment, $\varepsilon(0)$ noise variable for control, $\varepsilon(1)$ noise variable for treatment, and $Y$ the true outcome, all with length $n$.
\begin{equation}
    X \sim Uniform([0,1],size = (n,d)) 
\end{equation}
\begin{equation}
    b = sin(\pi  X_0  X_1)+2(X_2-0.5)^2+X_3+0.5X_4
\end{equation}
\begin{equation}
    \pi = max(0.1,min(sin(\pi  X_0  X_1),0.9)
\end{equation}
\begin{equation}
    \tau = \frac{X_0+X_1}{2}
\end{equation}
\begin{equation}
    W \sim Binomial(1, \pi)
\end{equation}
\begin{equation}
    \sigma_{het} = \sigma * \log(1 + |X_1|)
\end{equation}
\begin{equation}
    \varepsilon(0) \sim \mathcal{N}(0,\sigma^2)
\end{equation}
\begin{equation}
    \varepsilon(1) = c \varepsilon(0) + (1-c) \mathcal{N}(0,\sigma^2)
\end{equation}
\begin{equation}
    Y(0) = b-0.5\tau + \varepsilon(0)
\end{equation}
\begin{equation}
    Y(1) = b + 0.5\tau + \varepsilon(1)
\end{equation}
\begin{equation}
    Y = Y(0)(1-W)+ Y(1) W
\end{equation}
The models only use $Y$, $X$, $W$. All the other variables are for evaluation and testing purposes. 

\paragraph{Data generating process: Setup B} 
Setup B mimics a randomized trial. All other variables are calculated similarly as in setup A except for $X, b,\pi,$ and $\tau$.
\begin{equation}
    X \sim \mathcal{N}(0,1,size = (n,d)) 
\end{equation}
\begin{equation}
    b = max(0,X_0+X_1+X_2)+max(0,X_3+X_4)
\end{equation}
\begin{equation}
    \pi = 0.5
\end{equation}
\begin{equation}
    \tau = X_0+\log(1+exp(X_0))
\end{equation}

\paragraph{Data generating process: Setup C} 
Setup C mimics a situation with confounding with an easy propensity and a difficult baseline. All other variables are calculated similarly as in setup B except for $b$, $\pi$, and $\tau$.
\begin{equation}
    b = 2 \log(1+X_0+X_1+X_2)
\end{equation}
\begin{equation}
    \pi = \frac{1}{1+\exp(X_0+X_1+X_2)}
\end{equation}
\begin{equation}
    \tau = 1
\end{equation}

\paragraph{Data generating process: Setup D} 
Setup D mimics a situation with unrelated treatment and control groups. All other variables are calculated similarly as in setup B except for $b$, $\pi$, and $\tau$.
\begin{equation}
    b = \frac{max(0,X_0+X_1+X_2)+max(0,X_3+X_4)}{2}
\end{equation}
\begin{equation}
    \pi = \frac{1}{1+exp(-X_0)+exp(-X_1)}
\end{equation}
\begin{equation}
    \tau = max(0,X_0+X_1+X_2) - max(0,X_3+X_4)
\end{equation}

\subsection{Semi-synthetic experiments}
\label{sec:semi-exp}
The learners are also evaluated on semi-synthetic datasets. Four semi-synthetic datasets/experiments are used: IHDP~\citep{hillBayesianNonparametricModeling2011a}, NLSM~\citep{carvalhoAssessingTreatmentEffect2019}, EDU~\citep{zhou_estimating_2021}, and the ACIC 2016\citep{dorie_automated_2019}. The simulations run 100 times with $\alpha=0.1$, 100 MC samples, and with split WCPS using the residual as a conformity measure. Since these datasets do not provide propensity scores, we perform logistic regression to estimate them. For the ACIC 2016 experiments, we run the simulations 10 times because it has 77 different settings. The results are found in Figure~\ref{fig:semi-synthetic-dist}, \ref{fig:edu-ihdp-dist}, \ref{fig:pit-values}, \ref{fig:results_acic2016}, and \ref{fig:results_90_ihdp_nlsm}.

\subsubsection{IHDP}
The infant health and development program (IHDP) is a randomized controlled trial to evaluate the effect of home visits from physicians on the cognitive test scores of premature infants~\cite{hillBayesianNonparametricModeling2011a}. This dataset has 25 covariates with 6 continuous and 19 binary. Using this dataset, a semi-synthetic benchmark was created analogous to the study by \citet{hillBayesianNonparametricModeling2011a}, using setup B of their work. The used datasets are provided by \citet{shalitEstimatingIndividualTreatment2017}. The final dataset consists of 747 samples, 139 are treated samples and 608 of them are control samples. The data generation process of the semi-synthetic data is as follows:
\begin{equation}
    Y(0) \sim \mathcal{N}(exp(X+L)\beta,1)
\end{equation}
\begin{equation}
    Y(1)\sim \mathcal{N}(exp(X\beta)-\omega,1)
\end{equation}
The $\omega$ is always chosen such that ATT (Average Treatment effect on the Treated) is set to 4. The coefficients $\beta$ are chosen at random from 
\begin{equation*}
    (0,0.1,0.2,0.3,0.4)
\end{equation*}
each with a probability of 
\begin{equation*}
    (0.6,0.1,0.1,0.1,0.1)
\end{equation*}
respectively. $X$ represents the covariates in the IHDP dataset and $L$ is an offset matrix. The same treatment variable is kept as in the original IHDP study.

\subsubsection{NLSM}
The national study of learning mindsets (NLSM) is a large-scale randomized trial to investigate the effect of behavioural intervention on the academic performance of students~\citep{carvalhoAssessingTreatmentEffect2019}. Using the original NLSM dataset, new data is synthetically generated that follow the same distributions as the NLSM dataset as elaborated in \citet{carvalhoAssessingTreatmentEffect2019, alaaConformalMetalearnersPredictive2023}. The resulting dataset contains 10000 data points and 11 covariates ($X = [X_0, X_1, ...,X_{10}]$). The data generation process for the semi-synthetic data is as follows: 
\begin{equation}
    \gamma \sim \mathcal{N}(0,1,size=76)*0.105
\end{equation}
\begin{equation}
    \tau = 0.228 + 0.05\mathbb{I}(X_6 < 0.07) - 0.05\mathbb{I}(X_7 < -0.69) - 0.08\mathbb{I}(X_4 \in [1,13,14]) + \gamma_{X_0-1}
\end{equation}
with $\mathbb{I}$ the indicator function and $\gamma_{X_0-1}$ the $(X_0-1)$th element of an array of Gaussian samples with size 76. Then the outcome is defined as follows: 
\begin{equation}
    Y = W(Y-\tau)+(1-W)\tau
\end{equation}
with W as the treatment variable.

\subsubsection{EDU}
The EDU experimental setup, introduced by \citet{zhou_estimating_2022}, evaluates the impact of maternal adult education on children’s learning outcomes using a semi-synthetic dataset derived from a randomized field experiment in India \citep{banerji2017impact}. The binary treatment corresponds to maternal education receipt, and the outcome is the difference in children’s test scores. After preprocessing (sample size: 8,627; 32 covariates), synthetic potential outcomes were generated by training neural networks $\hat{f}_{Y(0)}(\cdot)$ (control) and $\hat{f}_{Y(1)}(\cdot)$ (treatment) on observed outcomes. Uncertainty was modeled as $Y_i(0) \sim \hat{f}_{Y(0)}(X_i) + (2 - m_i)\mathcal{N}(0, 0.5^2)$ and $Y_i(1) \sim \hat{f}_{Y(1)}(X_i) + (2 - m_i)\exp(2)$, where $m_i$ indicates prior maternal education, introducing heterogeneous variance. Treatment imbalance was induced via a high-coefficient logistic propensity model $\mathbb{P}(W_i=1|X_i) = (1 + \exp(-X_i^T\beta))^{-1}$ and truncation of subjects with propensities in $[0.3, 0.7]$. This design tests models under distributional heterogeneity and imbalance while aligning with real-world decision-making contexts.

\subsubsection{ACIC 2016}
The dataset from the 2016 Atlantic causal inference conference was created for a causal inference competition \citep{dorie_automated_2019}. It includes 4,802 observations with 58 covariates (3 categorical, 5 binary, 27 counts, and 23 continuous). The competition used these covariates to generate 77 different settings, each with varying outcome surface complexity, degrees of confounding, overlap, and treatment effect heterogeneity. We standardized all covariates and, for each setting, conducted 10 simulation runs, with a test set comprising 20\% of the data, randomly generated, following the methodology of \citet{curth_inductive_2021, alaaConformalMetalearnersPredictive2023}.
\end{document}